\documentclass{article}

\usepackage{microtype}
\usepackage{graphicx}
\usepackage{booktabs} 
\usepackage{multirow}
\usepackage{multicol}
\usepackage{tikz}
\usepackage[]{subfig}
\usepackage{adjustbox}
\usepackage{dblfloatfix}

\usepackage{hyperref}

\usepackage[accepted]{icml2023}

\usepackage{amsmath}
\usepackage{amssymb}
\usepackage{mathtools}
\usepackage{amsthm}

\DeclareMathOperator*{\argmin}{argmin} 

\newcommand{\li}{\mathcal{L}_i}
\newcommand{\lo}{\mathcal{L}_o}
\newcommand{\nin}{n_{\textrm{inner}}}
\newcommand{\nhes}{n_{H^{-1}}}
\newcommand{\nsub}{n_\textrm{subset}}

\usetikzlibrary{arrows.meta,
                calc,
                positioning,
                quotes}

\usepackage[capitalize,noabbrev]{cleveref}

\tikzstyle{graphnode} = [circle,text centered, draw=black, fill=red!30, minimum size = 1.2cm, inner sep=-3pt]
\tikzstyle{hiddennode} = [circle,text centered, draw=black, fill=yellow!30, minimum size = 1.2cm, inner sep=-3pt,]
\tikzstyle{endnode} = [circle,text centered, draw=black, fill=orange!30, minimum size = 1.2cm]
\tikzstyle{emptynode} = [circle,text centered, draw=none, fill=white!30, minimum size = 0.3cm, inner sep=-3pt,]
\tikzstyle{forwardarrow} = [thick,->,>=stealth, draw = {rgb:red,1;green,0;blue,3}, line width=0.5mm]
\tikzstyle{forwardarrow_faded} = [thick,->,>=stealth, draw = {rgb:red,1;green,0;blue,3}, line width=0.5mm, opacity = 0.2]
\tikzstyle{backwardarrow} = [thick,->,>=stealth, draw = {rgb:red,1;green,12;blue,1}, line width=0.5mm]

\usetikzlibrary{calc, decorations.pathreplacing,calligraphy}
\usetikzlibrary{arrows, arrows.meta}
\newlength{\nodevertical}
\newlength{\nodehorizontal}
\newlength{\nodehorizontallong}
\newlength{\fadeopacity}
\theoremstyle{plain}

\theoremstyle{definition}

\theoremstyle{remark}

\usepackage[textsize=tiny]{todonotes}

\icmltitlerunning{Dataset Distillation with Convexified Implicit Gradients}

\begin{document}

\twocolumn[
\icmltitle{Dataset Distillation with Convexified Implicit Gradients}




\begin{icmlauthorlist}
\icmlauthor{Noel Loo}{yyy}
\icmlauthor{Ramin Hasani}{yyy}
\icmlauthor{Mathias Lechner}{yyy}
\icmlauthor{Daniela Rus}{yyy}
\end{icmlauthorlist}

\icmlaffiliation{yyy}{Computer Science and Artificial Intelligence Lab (CSAIL), Massachusetts Institute of Technology (MIT)}

\icmlcorrespondingauthor{Noel Loo}{loo@mit.edu}

\icmlkeywords{Machine Learning, ICML}

\vskip 0.3in
]



\printAffiliationsAndNotice{\icmlEqualContribution} 

\begin{abstract}
We propose a new dataset distillation algorithm using reparameterization and convexification of implicit gradients (RCIG), that substantially improves the state-of-the-art. To this end, we first formulate dataset distillation as a bi-level optimization problem. Then, we show how implicit gradients can be effectively used to compute meta-gradient updates. We further equip the algorithm with a convexified approximation that corresponds to learning on top of a frozen finite-width neural tangent kernel. Finally, we improve bias in implicit gradients by parameterizing the neural network to enable analytical computation of final-layer parameters given the body parameters. RCIG establishes the new state-of-the-art on a diverse series of dataset distillation tasks. Notably, with one image per class, on resized ImageNet, RCIG sees on average a 108\% improvement over the previous state-of-the-art distillation algorithm. Similarly, we observed a 66\% gain over SOTA on Tiny-ImageNet and 37\% on CIFAR-100. \footnote{Code available at \url{https://github.com/yolky/RCIG}}
\end{abstract}

\section{Introduction}
Dataset distillation aims to condense a given dataset into a synthetic version that ideally preserves the information of the original set \citep{wang2018dataset}. Training on this synthetic set should result in similar performance compared to training on the original dataset.

Dataset distillation can be formulated as a bi-level optimization problem with an inner objective to update model parameters on the support/synthetic/distilled set, and an outer (meta) objective to refine the distilled sets via meta-gradient updates \citep{wang2018dataset}. Evaluating the meta-optimization loop is difficult, as we have to solve the inner optimization loop and then back-propagate errors through it to obtain the meta gradients. This could be done via backpropagation through time \citep{BPTT1}, evolution strategies \cite{evolution_strategies}, and equilibrium propagation \citep{equilibrium_prop}. Numerous research works have improved dataset distillation by introducing surrogate objectives for computing the meta gradients through gradient matching \citep{zhao2021dataset}, training trajectory matching \citep{cazenavette2022dataset}, feature alignment \citep{wang2022cafe}, by regressing on neural features \citep{frepo}, and by leveraging the neural tangent kernel (NTK) theory \citep{Jacot2018ntk, Arora_Exact_NTK_calc} in exact \citep{KIP1, KIP2} and approximate forms \citep{RFAD}. 

\emph{What does it take to significantly increase the accuracy and performance of dataset distillation?} In this paper, we construct a novel dataset distillation algorithm that substantially outperforms state-of-the-art methods, by re-parameterizing and convexifying implicit gradients. Implicit gradients (IGs) leverage implicit function theorem \citep{implicit_gradients1, implicit_gradients2, imaml}, that defines a meta-gradient update for a bi-level optimization problem (meta-learning with inner and outer loops). IG, off the bat, can serve as a dataset distillation algorithm, but it could perform significantly better via linearized training. Linearized training, which corresponds to learning on a frozen finite width neural tangent kernel (NTK) \cite{Jacot2018ntk}, convexifies the inner model, and as a result refines the implicit gradients. We show that such convexified implicit gradients dataset distillation algorithm equipped with a reparameterization technique for reducing the bias in IG considerably supersedes the state-of-the-art performance on 17 out of 22 reported benchmarks.

In summary, we make the following \textbf{new contributions:} 

\textbf{I.} We step-by-step construct a new dataset distillation algorithm called reparameterized convexified implicit gradients (RCIG), that establishes the new state-of-the-art. \textbf{II.} We show how to effectively design and improve implicit gradients to obtain a bi-level dataset distillation algorithm. 
\textbf{III.} We conduct a large experimental evaluation of our method in a diverse set of dataset distillation tasks and benchmarks and compare its performance to other advanced baselines.

\section{Background}

\textbf{Coresets and Dataset Distillation.}
Coresets are weighted subsets of the training data such that training on them results in the similar performance to training on the full dataset \cite{munteanu2018coresets, MirzasoleimanBL20,PooladzandiDM22}. 
Existing coreset selection methods employ clustering \cite{feldman2011unified, lucic2016strong, BachemLHK16}, bilevel optimization \cite{borsos2020coresets}, and sensitivity analysis \cite{munteanu2018coresets, HugginsCB16, TukanMF20}.

Dataset distillation shares many characteristics with coresets, however, instead of selecting subsets of the training data distillation generates synthetic samples. 
Similar to coresets, training on the synthetic samples should be faster and result in a better performing model \citep{wang2018dataset, zhao2021DC, zhao2021dsa, KIP1, KIP2, frepo, RFAD}.  Dataset distillation algorithms range from directly unrolling the model training computation graph \citep{wang2018dataset}, or approximately matching training trajectories with the full dataset \citep{mtt} As unrolling of the training comes with high memory and computation requirements, more recent works try to approximate the unrolled computation \citep{frepo, RFAD, KIP1, KIP2}. Dataset distillation has also shown promise in applications such as continual learning \citep{frepo, continual_dd}, and neural architecture search \citep{NAS_with_DD}.

\textbf{Bilevel Optimization and Implicit gradients.}
Bilevel optimization problems are a class of problems where one optimization problem is nested inside a second optimization problem. Formally, define the inner objective as $\mathcal{L}_{i}$ and the outer (meta) objective as $\mathcal{L}_o$. Bilevel optimization problems aim to solve:
\begin{align*}
    \argmin_{\psi} \mathcal{L}_o(\theta^*, \psi), \quad s.t. \quad \theta^* \in \argmin_\theta \mathcal{L}_i(\theta, \psi)
\end{align*}
With $\theta$ is a set of inner parameters, and $\psi$ is a set of outer/hyperparameters that we aim to solve for. This type of problem arises in many deep learning fields such as hyperparameters optimization \citep{bilevel_hyper1, bilevel_hyper2, bilevel_hyper3}, meta-learning \citep{maml, imaml}, and adversarial training \citep{OGadversarial, madry_pgd}. Similarly, dataset distillation can also be framed as a bilevel optimization problem, with $\theta$ the set of network parameters, and $\psi$ our distilled dataset parameters, given by the coreset images and labels \citep{wang2018dataset, KIP1, frepo}.

Evaluating $\mathcal{L}_o(\theta^*, \psi)$, and evaluating meta-gradients $\frac{d L_o}{d\psi}$ is difficult, as it generally not only requires solving the inner optimization problem to evaluate the outer loss but even worse, to backpropagate through inner optimization to compute meta-gradients. The most standard method is Backpropagation-through-time (BPTT) \citep{BPTT1, BPTT2, BPTT3}, which is what \citet{wang2018dataset} uses for dataset distillation. Other methods exist such as evolution strategies\citep{evolution_strategies, persistent_evolution} and equilibrium propagation \citep{equilibrium_prop, eqm_prop2}. One technique is \textit{implicit differentiation} (implicit gradients/IG methods), which leverages the implicit function theorem \citep{implicit_gradients1, implicit_gradients2, imaml}. This theorem states that if the inner object admits a unique minimum and the outer objective is continuously differentiable, then we have
\begin{align*}
    \frac{\partial \theta ^*}{\partial \psi} = \left(\frac{\partial^2 \li}{\partial \theta \partial \theta^T}\right)^{-1}\Bigg|_{\theta = \theta^*} \frac{\partial^2 \li}{\partial \theta \partial \psi}\Bigg|_{\theta = \theta^*}
\end{align*}
And our full meta-gradient is given by:
\begin{align}
    \frac{d\lo}{d\psi} = \underbrace{\frac{\partial \lo}{\partial \psi}}_{\textrm{direct grad}} + \underbrace{\frac{\partial}{\partial \psi}\left(\frac{\partial \li}{\partial \theta}^\intercal v\right)}_\textrm{implicit grad} \label{eq:full_meta_gradient}
\end{align}
With $v$ given by $\left(\frac{\partial^2 \li}{\partial \theta \partial \theta^T}\right)^{-1}\frac{\partial \lo}{\partial \theta}$.

\textbf{Neural Tangent Kernel and Network Linearization.}
The Neural Tangent Kernel (NTK) is a method of analyzing the behavior of neural networks in the infinite width limit \citep{Jacot2018ntk, arora2019on}. It states that as network width approaches infinity, with appropriate network initialization, neural networks behave as first-order Taylor expansion about their initialization, and are thus linear in their weights. The corresponding kernel formulation of the linear classifier results in the NTK, and the finite-width NTK converges to a deterministic architecture-dependent NTK in the infinite width limit. Networks behaving in this regime are said to be in the \textit{kernel regime}, and when the first-order Taylor expansion is accurate, networks are said to be in \textit{lazy training} \citep{lazy_training}. While typical neural networks have highly non-convex loss landscapes, networks in the NTK regime have \textbf{convex} loss landscapes. This convexity has been successfully exploited for tasks such as dataset distillation \citep{KIP1, KIP2, RFAD}, and federated learning \citep{TCT}.

\section{Method: Reparameterized Convexified Implicit Gradient}
In this section, we step-by-step motivate and explain how we build our reparameterized convexified Implicit Gradient (RCIG) dataset distillation algorithm.

\textbf{3.1. Implicit Gradients with Dataset Distillation.}
Dataset distillation can directly be framed as bilevel optimization by letting $\lo = \mathcal{L}_{T}(\theta)$, $\li = \mathcal{L}_{S(\psi)}(\theta)$, with $\mathcal{L}_T$ and $\mathcal{L}_{S(\psi)}$ being training losses of the full training set and coreset/support set, with $\psi = \{X_S, y_S\}$ our set of coreset images/labels.

This simple formulation leads directly to a straightforward dataset distillation algorithm, provided that we can compute $v = H_S^{-1}g_T$, with $H_S = \frac{\partial^2 \li}{\partial \theta \partial \theta^T}$ and $g_T = \frac{\partial \lo}{\partial \theta}$. This can be done using a Neumann series, conjugate gradient methods, or the methods we discuss in \cref{sec:hinv_computation}, but assume that this computation can be done for the time being. We call this naive implicit gradient algorithm VIG (Vanilla implicit gradients). We evaluate in this algorithm on MNIST, CIFAR-10, and CIFAR-100 distilling 1, 10, and 50 images per class (except on CIFAR-100) on a three-layer convolutional network, with results shown in \cref{tab:linearization_ablation} \citep{MNIST, cifar}. We additionally use small $L_2$ regularization so that the Hessian inverse is properly defined.

Immediately we see that implicit gradients perform poorly, sometimes performing \textit{worse} than random images, but why? As discussed in \citet{implicit_BLO}, there existing a unique minimizer is a necessary condition for the implicit gradient theorem to hold. In contrast, deep neural networks are highly non-convex, so we should not expect vanilla implicit gradients to work out of the box. Furthermore, the implicit gradient method relies heavily on computing $g_S^T H^{-1}_S g_T$, with $g_S = \frac{\partial \li}{ \partial \theta}$ (i.e. the support set gradient). This expression has a striking similarity to \textit{influence functions} \citep{old_influence, influence_fns, influence_fns_are_fragile}, which leads to a second interpretation of implicit-gradient based dataset distillation as maximizing the mutual influence between our distilled dataset and the full dataset.

While this interpretation of dataset distillation as influence maximization is appealing, it also suggests that the success of our algorithm is heavily dependent on how well $g_S H^{-1}_S g_T$ actually approximates influence. Recent work \citep{if_influence_is_answer} has shown that for deep models, these influence functions are brittle and do not accurately estimate leave-one-out retraining (which influence functions claim to do). \citet{if_influence_is_answer} shows that this discrepancy is partially caused by the non-convexity of deep models, as deep models undergo a period of highly erratic optimization behavior before only settling in approximately convex loss region. These two findings suggest that using implicit gradients to perform dataset distillation for highly non-convex deep networks is challenging.

\begin{table*}[t]
\caption{Performance of Vanilla Implicit Gradients (VIG), Convexified Implicit Gradients (CIG), and Reparameterization Covexified Implicit Gradients (RCIG), on distilling MNIST, CIFAR-10, and CIFAR-100. Linearization/convexification improves performance on almost all datasets, and reparameterization further improves performance to achieve state-of-the-art. (n=15)}
\centering
\resizebox{\textwidth}{!}{%
\begin{tabular}{lcccccccc}\toprule
              & \multicolumn{3}{c}{MNIST} & \multicolumn{3}{c}{CIFAR-10} & \multicolumn{2}{c}{CIFAR-100} \\ \cmidrule{2-9}
Img/cls       & 1      & 10      & 50     & 1       & 10       & 50      & 1             & 10            \\ \midrule
Random Subset &$62.73 \pm 0.95$	&	$93.22 \pm 0.16$	&	$97.79 \pm 0.12$	&	$20.76 \pm 0.49$	&	$38.43 \pm 0.36$	&	$54.44 \pm 0.34$	&	$6.24 \pm 0.08$	&	$21.08 \pm 0.11$	\\
VIG & $77.42 \pm 1.41$	&	$90.17 \pm 0.61$	&	$91.43 \pm 0.35$	&	$26.54 \pm 1.20$	&	$54.61 \pm 0.13$	&	$35.63 \pm 0.59$	&	$17.79 \pm 0.13$	&	$29.30 \pm 0.13$ \\
CIG (+ Linearization) & $69.23 \pm 1.43$	&	$95.37 \pm 0.26$	&	$95.78 \pm 0.17$	&	$29.70 \pm 0.95$	&	$56.48 \pm 0.60$	&	$56.68 \pm 0.57$	&	$19.72 \pm 0.55$	&	$31.36 \pm 0.23$ \\
RIG ( + Reparam) & $ \textbf{94.80} \pm 0.43 $ & $ 98.55 \pm 0.11 $ & $ 98.88 \pm 0.08 $ & $ 44.48 \pm 4.33 $ & $ 66.16 \pm 0.78 $ & $ 62.07 \pm 4.02 $ & $ 18.39 \pm 2.84 $ & $ \textbf{46.09} \pm 0.23 $ \\
RCIG (+ Lin + Reparam) & $\textbf{94.79} \pm 0.35$	&	$\textbf{98.93} \pm 0.03$	&	$\textbf{99.23} \pm 0.03$	&	$\textbf{52.75} \pm 0.76$	&	$\textbf{69.24} \pm 0.40$	&	$\textbf{73.34} \pm 0.29$	&	$\textbf{39.55} \pm 0.16$	&	$44.14 \pm 0.25$ \\ \bottomrule
\end{tabular}
}
\label{tab:linearization_ablation}
\end{table*}

\textbf{3.2. Convexification.} The literature and our simple experiments strongly suggest that our naive implicit gradients will not work unless we are able to make our inner model exhibit more convex behavior. One method of this is by considering the tangent space of the neural network parameters. Specifically, we define:
\begin{align}
    f_\theta(x) \approx f_{lin, \theta}(x) = f_{\theta_0}(x) + (\theta - \theta_0)^\intercal \nabla_\theta f_{\theta_0}(x).\label{eq:linearized_dyanmics}
\end{align}
We call this 1st-order Taylor approximation of learning dynamics \textit{linearized dynamics} \citep{fortman, loo2022evolution, wide_linear_models}, as opposed to standard dynamics which corresponds to no Taylor expansion. If we fix $\theta_0$ and consider optimizing $\theta$, this new formulation is now strongly convex in $\theta$, provided that some $L_2$ regularization is added. Seeing that at initialization, $f_{\theta_0}(x) \approx 0$,  we then have \textit{centered} linearized training. This can be efficiently calculated using forward-mode auto-differentiation.

This convex approximation, while fixing our non-convexity issue, is only as useful as it is accurate. This linearization technique corresponds to learning on top of a frozen finite-width neural tangent kernel (NTK) \citep{Jacot2018ntk, wide_linear_models, finite_dynamcis2, finite_vs_infinite, Hanin2020Finite}. For very wide neural nets, it has been shown that neural networks approximately behave as lazy/linear learners, with the correspondence increasing with wider networks \citep{wide_linear_models}. For narrow networks, it has been shown that networks undergo a brief period of rapid NTK evolution before behaving approximately as lazy learners \citep{fortman, loo2022evolution, Hanin2020Finite}.

This leads to a second implicit-gradient-based dataset distillation algorithm, which we call CIG (convexified implicit gradients). Now we replace both the inner and outer objectives from VIG with their convexified counterparts. We show CIG's performance in \cref{tab:linearization_ablation} and note that we evaluate using networks under standard dynamics (i.e. the evaluation network is unmodified). We see that convexification/linearization consistently improves distillation performance. While it is interesting that convexifying the problem improves distillation performance, the performance still falls short in state-of-the-art distillation algorithms, which can achieve over $65\%$ on CIFAR-10 with 10 images per class \citep{frepo}. Next, we close this gap by reparameterizing the problem to enable faster convergence.

\textbf{3.3. Combine Analytic Solutions and Implicit Gradients.}
\label{sec:rcig_description}
A key limitation of CIG is that in practice we are unable to find the true minima of the inner problem in any feasible amount of time. These truncated unrolls lead to bias in implicit gradients. Because the problem is convex, we can consider warm-starting the problem by reusing old intermediate values \citep{implicit_BLO}, but this still biases our results as warm-starts are only unbiased if we achieve the minimum every time, which we cannot feasibly do. From an algorithmic standpoint, warm starting also leads to a tradeoff in terms of model diversity, as we are forced to reuse the same model as opposed to instantiating new ones, which has shown to be vital in dataset distillation \citep{frepo, mtt}.

To perform better optimization of the inner problem, we exploit the structure of the neural network. Specifically, we split the network parameters into the final layer parameters, $\theta_F$ and the body parameters $\theta_B$, and note that for any setting of $\theta_B$, we can \textbf{efficiently analytically compute the optimal} $\mathbf{\theta_F}$ when trained under mean squared error (MSE) loss. Specifically, consider MSE loss with labels $y_S$, and let $h_{\theta_0}(X_S) \in R^{H\times |S|}$ be the hidden layer embeddings, with $|S|$ the distilled dataset size and $H$ the final layer dimension. We know that $\nabla_{\theta}{f_{lin, \theta} = h_{\theta_0}(X_S)}$. Defining $k^{\theta_0} (x, x') = h_{\theta_0}(x)^T h_{\theta_0}(x')$, be the associated final-layer NTK, neural network gaussian process (NNGP) or conjugate kernel, we have the optimal set of final layer parameters for our centered linearized problem is given by
\begin{align}
    \theta_F^* &= h_{\theta_0}(X_S) \left(K^{\theta_0}_{X_S, X_S} + \lambda I_{|S|}\right)^{-1}\hat{y}_S  \label{eq:optimal_theta_f} \\
    \hat{y}_S\ &= \left(y_S - \theta_B ^\intercal \frac{\partial f_{lin, \theta}(X_S)}{\partial \theta_B}\right), \label{eq:perturbed_labels}
\end{align}
where $\theta_B ^\intercal \frac{\partial f_{lin, \theta}(X_S)}{\partial \theta_B}$ could be considered an \emph{offset} which changes the labels given by how much the body parameters already changed. Note that without this offset, this method of solving the optimal final layer parameters corresponds to training using the NNGP/Conjugate kernel to convergence, which has been used in \citet{RFAD} and \citet{frepo}. However, these methods ignore the contribution from the body parameters, which our method does not.

Given that we can solve for the optimal $\theta_F$ given $\theta_B$, we now reparameterize our problem so that we \textbf{only learn} $\mathbf{\theta_B}$, and automatically compute $\theta_F$. Specifically, our parameterized inner and outer objectives then become:
\begin{align}
\label{eq:inner_loss}
    \li^{\textrm{rep}}(\theta_B) = \mathcal{L}_{S(\psi)}(\theta_B, \theta_F^*(\theta_B, \psi) ),\hspace{0.5em} \text{and}
    \end{align}
\begin{align}
\label{eq:outer_loss}
    \lo^{\textrm{rep}}(\theta_B, \psi) &= \mathcal{L}_{platt, T}(\theta_B, \theta_F^*(\theta_B, \psi), \tau).
\end{align}

We additionally add $L_2$ regularization to the inner objective $\frac{\lambda}{2}(\theta_B^\intercal \theta_B + \theta_F^*(\theta_B)^\intercal \theta_F^*(\theta_B))$. For the outer objective, we adopt the same Platt scaling loss with a learnable temperature parameter $\tau$ used in \citep{RFAD}, as it has shown to be highly effective in dataset distillation settings. At a high level, the Platt scaling loss replaces the MSE loss with $\mathcal{L}_{platt} = \textrm{xent}(\hat{y}/\tau, y)$, with $\hat{y}$ our predictions, $\tau$ the learnable temperature parameter and $y$ the true labels and $\textrm{xent}$ the cross entropy function.

Thus, our final meta-gradient is:
\begin{align*}
    \frac{d\lo^\textrm{rep}}{d\psi} = \underbrace{\frac{\partial \lo^\textrm{rep}}{\partial \psi}}_{\textrm{direct grad}} + \underbrace{\frac{\partial}{\partial \psi}\left(\frac{\partial \li^\textrm{rep}}{\partial \theta_B}^\intercal v\right)}_\textrm{implicit grad},
\end{align*}
with $v = H_S^{\textrm{rep}, -1}g_T^{\textrm{rep}}$, with $H^\textrm{rep}_S = \frac{\partial^2 \li^\textrm{rep}}{\partial \theta_B \partial \theta_B^T}$ and $g_T^\textrm{rep} = \frac{\partial \lo^\textrm{rep}}{\partial \theta_B}$. Unlike CIG, this reparameterized version has a non-zero contribution from the direct gradient, as $\psi$ is used to compute the optimal set of $\theta_F$. When using an MSE-loss as opposed to the Platt-loss, this direct gradient corresponds to the FRePo \citep{frepo} loss, evaluated using the perturbed labels given in \cref{eq:perturbed_labels}, thus we could consider our algorithm to be a generalization of FRePo to consider the body parameters as well as the final layer parameters.

Finally, noting that neural networks undergo a period of rapid kernel evolution early in training \citep{fortman, Hanin2020Finite, loo2022evolution}, it is important to not only use the initialization finite-width NTK but also the evolved NTK. Thus we adopt the same technique used in \citet{frepo}, where we have a pool of partially trained models. We fix this pool to contain $m = 30$ models and set the max number of training steps of these models to be $K = 100$, in line with \citet{frepo}. Next, we discuss computing $v = H_S^{\textrm{rep}, -1}g_T^{\textrm{rep}}$.

\textbf{3.4. Hessian-inverse Vector Computation.}
\label{sec:hinv_computation}
To compute implicit gradients, we need to compute $v = H_S^{\textrm{rep}, -1}g_T^{\textrm{rep}}$. As $H$ has dimension $P \times P$, where $P$ is the parameter dimension, this cannot be done exactly. Thus, it is typical to approximate $v$ using methods such as conjugate gradients \citep{CG_in_deep_learning} or the Neumann series \citep{neumann_unroll, influence_fns}. The method we use is closely related to the Neumann series. We note that $v$ is a minimizer of the following loss:
\begin{align}
    \mathcal{L}_{H^{-1}} = (Hv - g)^\intercal H^{-1} (Hv - g),\label{eq:hinv_loss}
\end{align}
which has gradients w.r.t $v$ as $Hv - g$. Thus we perform stochastic gradient descent (SGD) on this objective using an optimizer such as Adam for $\nhes$ gradient steps. Note that using SGD on this objective corresponds to the Neumann series method of computing Hessian-inverse vector products. Hessian-vector products can be efficiently computed using the Perlmutter trick \citep{perlmutter}. For the inner optimization objective, we perform $\nin$ optimization steps and then $\nhes$ optimization steps on \cref{eq:hinv_loss}. This leads to the Reparameterized Convexified Implicit Gradients (RCIG) algorithm, with pseudo-code given in \cref{alg:rcig}.

\textbf{Complexity Analysis.} Let our coreset size be $|S|$, training batch size be $|B|$, and our network parameter have dimension $P$. One training iteration, per \cref{alg:rcig} contains three main steps: optimizing the inner objective, computing the Hessian-inverse vector product, and computing the meta-gradient. Optimizing the inner objective takes $O(\nin|S|)$ time, as it requires a full forward pass of the coreset at each training iteration to compute \cref{eq:inner_loss}. Likewise, computing the Hessian-inverse vector product takes $O(\nhes |S|)$ time, as we perform $\nhes$ optimization steps on \cref{eq:hinv_loss}. Computing the direct gradient and implicit gradient also costs $O(|S|)$ time, giving a total time complexity of $O((\nin + \nhes)|S|)$. The memory requirements of inner optimization and Hessian inverse computation are constant in $\nin$ and $\nhes$, as we do not retain intermediate computations, so the total memory consumption is $O(P)$.
\vspace{0.2in}

\newcommand{\ahes}{\alpha_{H^{-1}}}
\newcommand{\ain}{\alpha_{\textrm{inner}}}
\newcommand{\acs}{\alpha_S}

\begin{algorithm}[t]
   \caption{Reparam Convexified Implicit Gradients}
   \label{alg:rcig}
\begin{algorithmic}
   \STATE {\bfseries Input:} Training set and labels $\mathcal{T}$, inner, Hessian-inverse and distilled dataset learning rates $\ain, \ahes, \acs$\\
   \STATE {\bfseries Initialize:} Initialize distilled dataset and labels $\mathcal{S}$ with parameters $\psi = \{X_S, y_S, \tau \}$
   \STATE {\bfseries Initialize:} Initialize a model pool $\mathcal{M}$ with m randomly initialized models $\{ \theta_i \}_{i = 1}^m$
   \WHILE{not converged}
    \STATE Sample a random model from the pool: $\theta_i \sim \mathcal{M}$
    \STATE Sample a training batch from the training set: $\{X_T, y_T\} \sim \mathcal{T}$
    \STATE Perform $\nin$ optimization steps on inner objective $\li^\textrm{rep}$ given by \cref{eq:inner_loss} to obtain $\theta^*_i$ with \textbf{linearized} dynamics
    \STATE Perform $\nhes$ optimization steps on \cref{eq:hinv_loss} to obtain $v$
    \STATE Compute direct gradient $g_\textrm{direct} = \frac{\partial \lo^\textrm{rep}}{\partial \psi}$, with $\lo^\textrm{rep}$ given by \cref{eq:outer_loss}
    \STATE Compute implicit gradient $g_\textrm{implicit} = \frac{\partial}{\partial \psi}\left(\frac{\partial \li^\textrm{rep}}{\partial \theta_B}^\intercal v\right)$
    \STATE Update the distilled dataset: $\psi \leftarrow \psi - \acs (g_\textrm{direct} + g_\textrm{implicit})$
    \STATE Train the model $\theta_i$ on the current distilled dataset $\mathcal{S}$ for one step using \textbf{standard} dynamics
    \STATE Reinitialize the model $\theta_i$ if it has been updated for more than $K$ steps.
   \ENDWHILE
\end{algorithmic}
\label{alg:rcig}
\end{algorithm}

\textbf{3.6. Bias-free subsampling.}
\label{sec:subsampling}
One limitation of RCIG compared to CIG is that when optimizing the inner objective, we need to compute $\nabla_{\theta_B} (\mathcal{L}_{S(\psi)}(\theta_B, \theta_F^*(\theta_B))$, which we know from \cref{eq:optimal_theta_f}, depends on the entire training set $X_S, y_S$. Thus, we cannot use stochastic gradient descent as we can with CIG, as leaving out elements of $X_S$ would lead to biased gradients. Likewise, when computing $H^{\textrm{rep}, -1}_S g_T^\textrm{rep}$, we need to compute Hessian-vector products, which again rely on all of $X_S, y_S$. This, combined with the fact that linearization, in general, incurs a doubling of memory cost, makes direct implementation of RCIG difficult for very large support sets (for example CIFAR-100 with 50 images per class). Here, we present a simple technique for getting unbiased gradients without incurring the full memory cost.

When computing these gradients, we note that the gradient contribution from each of the $X_S$ elements is interchangeable, implying that when performing the backward pass, the gradients computing through the nodes associated with $h_{\theta_0}(x_i)$, are all unbiased estimates of the gradient when computed through all nodes $h_{\theta_0}(X_S)$. This means that rather than computing the backward pass through all nodes, we can compute a backward pass through only some of the $h_{\theta_0}(X_S)$ nodes, provided that prior to those nodes the backward pass, the whole computation is stored. Note that from a memory perspective, computing $h_{\theta_0}(X_S)$ \textbf{is not expensive}, but requires storing the computation graph leading to that point for the backward pass \textbf{is expensive}. Thus, by randomly dropping parts of this computation graph we can save substantial memory. The important note is that during the backward pass, we still need accurate gradients leading into $h_{\theta_0}$, meaning that we still need to full forward pass so that we can compute $\theta_F^*$. 

Thus, we propose running a full forward pass on the whole support set, but stop gradients for a random subset of $h_{\theta_0}(X_S)$, only allowing the backward pass through the complementary subset of size $\nsub < |S|$, with $|S|$ the distilled dataset size. A more detailed description and formal justification for this technique are present in \citet{randomized_autodiff}. Schematically, this is shown in \cref{fig:subsampling}. We use this technique whenever the coreset size exceeds 1000 images.

\section{Experimental Results}
In this section, we present our comprehensive experimental evaluation of our method, RCIG, compared to modern baselines using a diverse series of benchmarks and tasks.

\begin{table*}[ht]
\centering
\caption{Distillation performance of RCIG and six baseline distillation algorithms on six benchmark datasets. RCIG attains the highest accuracy on 13/16 of these benchmarks, with the largest gains in the 1 Img/Cls category. (n=15)}
 \begin{adjustbox}{width=0.8\textwidth}
\begin{tabular}{lc|cccccc|c| c} \toprule
                            & Img/Cls & DSA & DM & KIP & RFAD & MTT & FRePo & RCIG  & Full Dataset\\ \midrule
\multirow{3}{*}{MNIST} & 1	&	$ 88.7 \pm 0.6 $	&	$ 89.9 \pm 0.8 $	&	$ 90.1 \pm 0.1 $	&	$ 94.4 \pm 1.5 $	&	$ 91.4 \pm 0.9 $	&	$ 93.0 \pm 0.4 $	&	$ \mathbf{94.7 \pm 0.5} $\\
                     & 10	&	$  97.9 \pm 0.1 $	&	$ 97.6 \pm 0.1 $	&	$ 97.5 \pm 0.0 $	&	$ 98.5 \pm 0.1 $	&	$ 97.3 \pm 0.1 $	&	$ 98.6 \pm 0.1 $	&	$ \mathbf{98.9 \pm 0.0 }$ & $99.6\pm 0.0$\\
                     & 50	&	$  99.2 \pm 0.1 $	&	$ 98.6 \pm 0.1 $	&	$ 98.3 \pm 0.1 $	&	$ 98.8 \pm 0.1 $	&	$ 98.5 \pm 0.1 $	&	$ \mathbf{99.2 \pm 0.0} $	&	$ \mathbf{99.2 \pm 0.0 }$\\ \midrule
\multirow{3}{*}{F-MNIST} & 1	&	$ 70.6 \pm 0.6 $	&	$ 71.5 \pm 0.5 $	&	$ 73.5 \pm 0.5 $	&	$ 78.6 \pm 1.3 $	&	$ 75.1 \pm 0.9 $	&	$ 75.6 \pm 0.3 $	&	$ \mathbf{79.8 \pm 1.1 } $\\
                     & 10	&	$  84.8 \pm 0.3 $	&	$ 83.6 \pm 0.2 $	&	$ 86.8 \pm 0.1 $	&	$ 87.0 \pm 0.5 $	&	$ 87.2 \pm 0.3 $	&	$ 86.2 \pm 0.2 $	&	$ \mathbf{88.5 \pm 0.2 }$ & $93.5 \pm 0.1$\\
                     & 50	&	$  88.8 \pm 0.2 $	&	$ 88.2 \pm 0.1 $	&	$ 88.0 \pm 0.1 $	&	$ 88.8 \pm 0.4 $	&	$ 88.3 \pm 0.1 $	&	$ 89.6 \pm 0.1 $	&	$ \mathbf{90.2 \pm 0.2 }$\\ \midrule
\multirow{3}{*}{CIFAR-10} & 1	&	$ 36.7 \pm 0.8 $	&	$ 31.0 \pm 0.6 $	&	$ 49.9 \pm 0.2 $	&	$ 53.6 \pm 1.2 $	&	$ 46.3 \pm 0.8 $	&	$ 46.8 \pm 0.7 $	&	$ \mathbf{53.9 \pm 1.0 }$\\
                     & 10	&	$  53.2 \pm 0.8 $	&	$ 49.2 \pm 0.8 $	&	$ 62.7 \pm 0.3 $	&	$ 66.3 \pm 0.5 $	&	$ 65.3 \pm 0.7 $	&	$ 65.5 \pm 0.4 $	&	$ \mathbf{69.1 \pm 0.4 }$ & $84.8\pm 0.1$\\
                     & 50	&	$  66.8 \pm 0.4 $	&	$ 63.7 \pm 0.5 $	&	$ 68.6 \pm 0.2 $	&	$ 71.1 \pm 0.4 $	&	$ 71.6 \pm 0.2 $	&	$ 71.7 \pm 0.2 $	&	$ \mathbf{73.5 \pm 0.3 }$\\ \midrule
\multirow{3}{*}{CIFAR-100} & 1	&	$ 16.8 \pm 0.2 $	&	$ 12.2 \pm 0.4 $	&	$ 15.7 \pm 0.2 $	&	$ 26.3 \pm 1.1 $	&	$ 24.3 \pm 0.3 $	&	$ 28.7 \pm 0.1 $	&	$ \mathbf{39.3 \pm 0.4 }$\\
                     & 10	&	$  32.3 \pm 0.3 $	&	$ 29.7 \pm 0.3 $	&	$ 28.3 \pm 0.1 $	&	$ 33.0 \pm 0.3 $	&	$ 40.1 \pm 0.4 $	&	$ 42.5 \pm 0.2 $	&	$ \mathbf{44.1 \pm 0.4 }$ & $56.2\pm0.3$\\
                     & 50	&	$  42.8 \pm 0.4 $	&	$ 43.6 \pm 0.4 $	&	-	&	-	&	$ \mathbf{47.7 \pm 0.2} $	&	$ 44.3 \pm 0.2 $	&	$ 46.7 \pm 0.3 $\\ \midrule 
\multirow{2}{*}{T-ImageNet} & 1	&	$ 6.6 \pm 0.2 $	&	$ 3.9 \pm 0.2 $	&	-	&	-	&	$ 8.8 \pm 0.3 $	&	$ 15.4 \pm 0.3 $	&	$ \mathbf{25.6 \pm 0.3 }$ & \multirow{2}{*}{$37.6\pm0.4$}\\
                     & 10	&	-	&	$ 12.9 \pm 0.4 $	&	-	&	-	&	$ 23.2 \pm 0.2 $	&	$ 25.4 \pm 0.2 $	&	$ \mathbf{29.4 \pm 0.2 }$\\ \midrule
\multirow{2}{*}{CUB-200} & 1	&	$ 1.3 \pm 0.1 $	&	$ 1.6 \pm 0.1 $	&	-	&	-	&	$ 2.2 \pm 0.1 $	&	$ \mathbf{12.4 \pm 0.2} $	&	$ 12.1 \pm 0.2 $ & \multirow{2}{*}{$20.5\pm0.3$}\\
                     & 10	&	$  4.5 \pm 0.3 $	&	$ 4.4 \pm 0.2 $	&	-	&	-	&	-	&	$ \mathbf{16.8 \pm 0.1 }$	&	$ 15.7 \pm 0.3 $\\ \bottomrule
\end{tabular}
\end{adjustbox}
\label{tab:main_res_table}
\end{table*}

\textbf{4.1. Results on Standard Benchmarks.}
\label{sec:standard_results}
We first ran RCIG on six standard benchmarks tests including MNIST (10 classes) \citep{MNIST}, Fashion-MNIST (10 classes) \citep{FASHION_MNIST}, CIFAR-10 (10 classes), CIFAR-100 (10 classes) \citep{cifar}, Tiny-ImageNet (200 classes) \citep{tiny_imagenet}, and Caltech Birds 2011 (200 classes) \citep{CUB-200}, with performances reported in \cref{tab:main_res_table}. We compare RCIG to six baseline dataset distillation algorithms including Differentiable Siamese Augmentation (DSA) \citep{zhao2021dsa}, Distribution Matching (DM) \citep{zhao2021dataset}, Kernel-Inducing-Points (KIP) \citep{KIP2}, Random Feature Approximation (RFAD) \citep{RFAD}, Matching Training Trajectories (MTT) \citep{mtt}, and neural Feature Regression with Pooling (FRePo) \citep{frepo}. 

Our method establishes the new state-of-the-art performance in 13 out of 16 of these benchmark tasks, sometimes with a significant margin. We see the greatest performance gains in datasets that have many classes, with few images per class. In CIFAR-100 with one image per class, we are able to achieve $39.3 \pm 0.4\%$, compared to the previous state of the art $28.7 \pm 0.4\%$, which is equivalent to \textbf{37\% improvement over SOTA}. Similarly, in Tiny-ImageNet with one image per class, we achieve $25.6 \pm 0.3\%$ compared to the previous state-of-the-art $15.4 \pm 0.3\%$, which is an improvement of \textbf{66\% over SOTA}. For CUB-200, we notice that we underperform compared to FRePo. Noting the relatively small training set size (5,994), compared to the number of classes (200), we observed that our algorithm tended to overfit to the training set, as with 10 images per class, we observed that we would achieve 100\% training classification accuracy. A solution to this would have been to apply data augmentation during dataset distillation, however, to keep our method simple we applied no augmentation. 

RCIG outperforms SOTA methods even with larger support sizes with a good margin. However, the performance gain is not as significant as in the case of smaller support sets. For example, we see a performance gain of 14\% going from 1 to 10 classes in Tiny-ImageNet over SOTA. We hypothesize that this is because minimizing the inner objective is harder when the dataset is larger, and likely would require a lower learning rate and more inner steps to converge.


\begin{table*}[t]
\caption{Performance of CIFAR-10 10 Img/Cls distilled datasets evaluated on different network architectures. Default Normalization (DN) refers to the test architecture used in reported results in the respective papers. RCIG-distilled datasets achieve high accuracy across a variety of datasets, particularly when BatchNorm (BN) is used during training. * - see footnote\footnote{We reevaluated the publicly available FrePo distilled dataset using the publicly available code to get these results. Additionally, we fixed an error in the code (see \cref{app:frepo_bug})} (n=15)}
\centering
 \begin{adjustbox}{width=0.7\textwidth}
\begin{tabular}{llcccccc} \toprule
              & \multirow{2}{*}{\begin{tabular}{@{}c@{}}Training \\ Architecture\end{tabular}} & \multicolumn{5}{c}{Evaluation Architecture} & \\ \cmidrule{3-8}
              &                                & Conv-DN & Conv-NN & ResNet-DN & ResNet-BN & VGG-BN & AlexNet \\ \midrule
DSA	&	Conv-IN	&	$53.2 \pm 0.8$	&	$36.4 \pm 1.5$	&	$42.1 \pm 0.7$	&	$34.1 \pm 1.4$	&	$46.3 \pm 1.3$	&	$34.0 \pm 2.3$	\\
DM	&	Conv-IN	&	$49.2 \pm 0.8$	&	$35.2 \pm 0.5$	&	$36.8 \pm 1.2$	&	$35.5 \pm 1.3$	&	$41.2 \pm 1.8$	&	$34.9 \pm 1.1$	\\
MTT	&	Conv-IN	&	$64.4 \pm 0.9$	&	$41.6 \pm 1.3$	&	$49.2 \pm 1.1$	&	$42.9 \pm 1.5$	&	$46.6 \pm 2.0$	&	$34.2 \pm 2.6$	\\
KIP	&	Conv-NTK	&	$62.7 \pm 0.3$	&	$58.2 \pm 0.4$	&	$49.0 \pm 1.2$	&	$45.8 \pm 1.4$	&	$30.1 \pm 1.5$	&	$57.2 \pm 0.4$	\\
FRePo	&	Conv-BN	&	$65.5 \pm 0.4$	&	$65.5 \pm 0.4$	&	$\mathbf{54.4 \pm 0.8}^*$	&	$52.4 \pm 0.7^*$	&	$\mathbf{55.4 \pm 0.7^*}$	&	$61.6 \pm 0.2^*$	\\
RCIG	&	Conv-NN	&	$\mathbf{69.1 \pm 0.4}$	&	$\mathbf{69.1 \pm 0.4}$	&	$51.3 \pm 1.7$	&	$49.7 \pm 1.4$	&	$46.2 \pm 1.8$	&	$60.8 \pm 1.8$	\\
RCIG	&	Conv-BN	&	$66.0 \pm 0.6$	&	$66.0 \pm 0.6$	&	$\mathbf{54.4 \pm 1.1}$	&	$\mathbf{54.8 \pm 1.1}$	&	$\mathbf{55.4 \pm 1.1}$	&	$\mathbf{62.1 \pm 0.8}$	\\ \bottomrule
\end{tabular}
\end{adjustbox}
\label{tab:arch_transfer}
\end{table*}

\textbf{4.2. Cross-architecture generalization.}
A desirable property of distilled datasets is the ability to transfer well to unseen training architectures. Here we evaluate the transferability of RCIG's distilled datasets for CIFAR-10 10 images per class. Following prior work \citep{zhao2021dsa, zhao2021dataset, mtt, frepo}, we evaluate our models on the ResNet-18 \citep{resnet}, VGG11 \citep{vgg}, and AlexNet \citep{alexnet}. Additionally, we consider various normalization layers such as using no normalization (NN), Batch Normalization (BN) \citep{batchnorm}, and Instance Normalization (IN). Default normalization (DN) refers to the evaluation architecture used in the respective paper, which is the same as the training architecture, except for FRePo, which trains with BN and evaluates with NN. RCIG Typically trains with NN and evaluates with NN, but we also consider training with BN. \cref{tab:arch_transfer} summarizes the results. We see that RCIG can achieve high transferability, in particular when we use BN during training. We hypothesize that using BN during training helps ensure that magnitudes of distilled images remain similar to real images, allowing for wider generalizability, although future work should investigate the role of normalization during training further.

\begin{table*}[t]
\centering
\caption{Distillation performance for ImageNet subsets. RCIG attains the highest performance in the single image per class category on all benchmarks. In particular, RCIG \textbf{doubles} the performance of the state-of-the-art in the ImageNet 1 Img/Cls setting. (n=15)}
\label{tab:imagenet_acc_table}
\begin{adjustbox}{width=0.7\textwidth}
\begin{tabular}{lcccccc} \toprule
              & \multicolumn{2}{c}{ImageNette (128x128)} & \multicolumn{2}{c}{ImageWoof (128x128)} & \multicolumn{2}{c}{ImageNet (64x64)} \\ \cmidrule(lr){2-3} \cmidrule(lr){4-5} \cmidrule(lr){6-7}
Img/Cls       & 1                   & 10                 & 1                   & 10                & 1                 & 2                 \\ \midrule
Random Subset & $ 23.5\pm 4.8$      & $47.4\pm 2.4$      & $14.2\pm 0.9$       & $27.0\pm 1.9$      & $1.1\pm0.1$       & $1.4\pm 0.1$       \\
MTT           & $47.7\pm 0.9$       & $63.0\pm 1.3$      & $28.6\pm 0.8$       & $35.8\pm 1.8$      & -                 & -                 \\
FRePo         & $48.1\pm 0.7$       & $\mathbf{66.5\pm 0.8}$      & $29.7\pm 0.6$       & $\mathbf{42.2\pm 0.9}$      & $7.5\pm0.3$       & $9.7\pm 0.2$       \\ \midrule
RCIG          &   $ \mathbf{53.0 \pm 0.9} $	 & 	$ 65.0 \pm 0.7 $	 & 	$ \mathbf{33.9 \pm 0.6} $	 & 	$ \mathbf{42.2 \pm 0.7} $	 & 	$ \mathbf{15.6 \pm 0.2} $	 & 	$ \mathbf{16.6 \pm 0.1} $ \\ \bottomrule
\end{tabular}
\end{adjustbox}
\end{table*}

\textbf{4.3. Experiments with ImageNet Datasets.}
We next considered higher-resolution subsets. Consistent with \citet{frepo} and \citet{mtt}, we consider two ImageNet subsets: ImageNette and ImageWoof, both subsets of ImageNet with 10 classes each with resolutions of $128\times 128$ \citep{imagenette}. On these two datasets, we see that RCIG outperforms the baselines substantially in the 1 image per class setting, but only performs similarly to the previous state-of-the-art with more images. To evaluate how well RCIG scales to more complex label spaces, we also consider the full ImageNet-1K dataset with 1000 classes, resized to $64 \times 64$ \citep{imagenet}. With one image per class, \textbf{our algorithm doubles the performance of the previous SOTA}, achieving $\mathbf{15.6\pm 0.2\%}$. 

\textbf{4.4. Ablation: Number of Optimization Steps.}
To run RCIG, we require two key hyperparameters, the number of inner optimization steps, $\nin$, and the number of steps used to compute the Hessian-inverse-vector product, $\nhes$. These two hyperparameters play a critical role in the fidelity of all implicit gradient methods. $\nin$ controls the quality of the inner optimization procedure, with larger $\nin$ resulting in better approximations of the true minimum. Likewise, $\nhes$ controls the accuracy of our approximation of $v = H^{-1}g_T$. In previous sections, we fixed $\nin = \nhes = 20$ for all datasets/experiments for simplicity. In this section, we study the effect of both these critical parameters on the runtime and accuracy of the algorithm.

\begin{figure}[b!]
\vskip 0.0in
\begin{center}
\includegraphics[width = 0.47\textwidth]{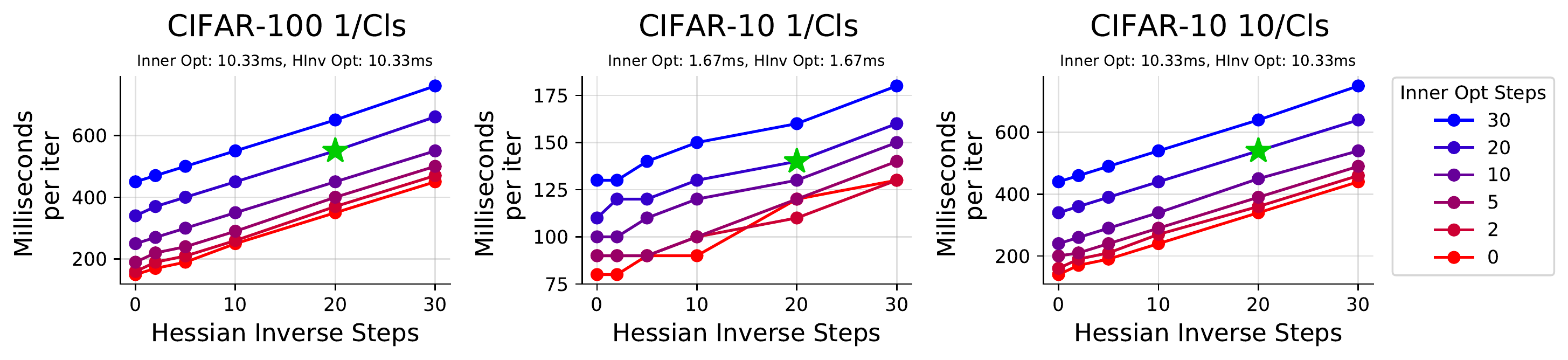}
\caption{The effect of $\nin$ and $\nhes$ on the computation time for CIFAR-100 1 Img/Cls, and CIFAR-10 with 1 and 10 Img/Cls. The green star denotes the hyperparameter configuration used throughout the paper ($\nin = 20, \nhes = 20$). 
}
\label{fig:cg_ablation_time_per_iter}
\end{center}
\vskip -0.2in
\end{figure}

\begin{table*}[t]
\centering
\caption{AUC of five MIA attack strategies for neural networks trained on distilled data. 
(n=25)}
\begin{adjustbox}{width=0.7\textwidth}
\begin{tabular}{lcccccc} \toprule
              & \multirow{2}{*}{Test Acc (\%)} & \multicolumn{5}{c}{Attack AUC}  \\ \cmidrule{3-7}
              &                                & Threshold & LR & MLP & RF & KNN \\ \midrule
Real	&	$\mathbf{99.2 \pm 0.1}$	&	$0.99 \pm 0.01$	&	$0.99 \pm 0.00$	&	$1.00 \pm 0.00$	&	$1.00 \pm 0.00$	&	$0.97 \pm 0.00$	\\
Subset	&	$96.8 \pm 0.2$	&	$0.52 \pm 0.00$	&	$\mathbf{0.50 \pm 0.01}$	&	$\mathbf{0.53 \pm 0.01}$	&	$0.55 \pm 0.00$	&	$0.54 \pm 0.00$	\\
DSA	&	$98.5 \pm 0.1$	&	$0.50 \pm 0.00$	&	$0.51 \pm 0.00$	&	$0.54 \pm 0.00$	&	$0.54 \pm 0.01$	&	$0.54 \pm 0.01$	\\
DM	&	$98.3 \pm 0.0$	&	$0.50 \pm 0.00$	&	$0.51 \pm 0.01$	&	$0.54 \pm 0.01$	&	$0.54 \pm 0.01$	&	$0.53 \pm 0.01$	\\
FRePo	&	$98.5 \pm 0.1$	&	$0.52 \pm 0.00$	&	$0.51 \pm 0.00 $	&	$\mathbf{0.53 \pm 0.01} $	&	$\mathbf{0.52 \pm 0.01} $	&	$\mathbf{0.51 \pm 0.01}$	\\
RCIG	&	$\underline{98.9 \pm 0.0}$	&	$\mathbf{0.49 \pm 0.00}$	&	$\mathbf{0.50 \pm 0.00}$	&	$\mathbf{0.53 \pm 0.00}$	&	$0.53 \pm 0.00$	&	$0.52 \pm 0.00$	\\ \bottomrule
\end{tabular}
\end{adjustbox}
\label{tab:mia_results_table}
\end{table*}

\begin{figure*}[t]
\vskip 0.0in
\begin{center}
\includegraphics[width = 0.7\linewidth]{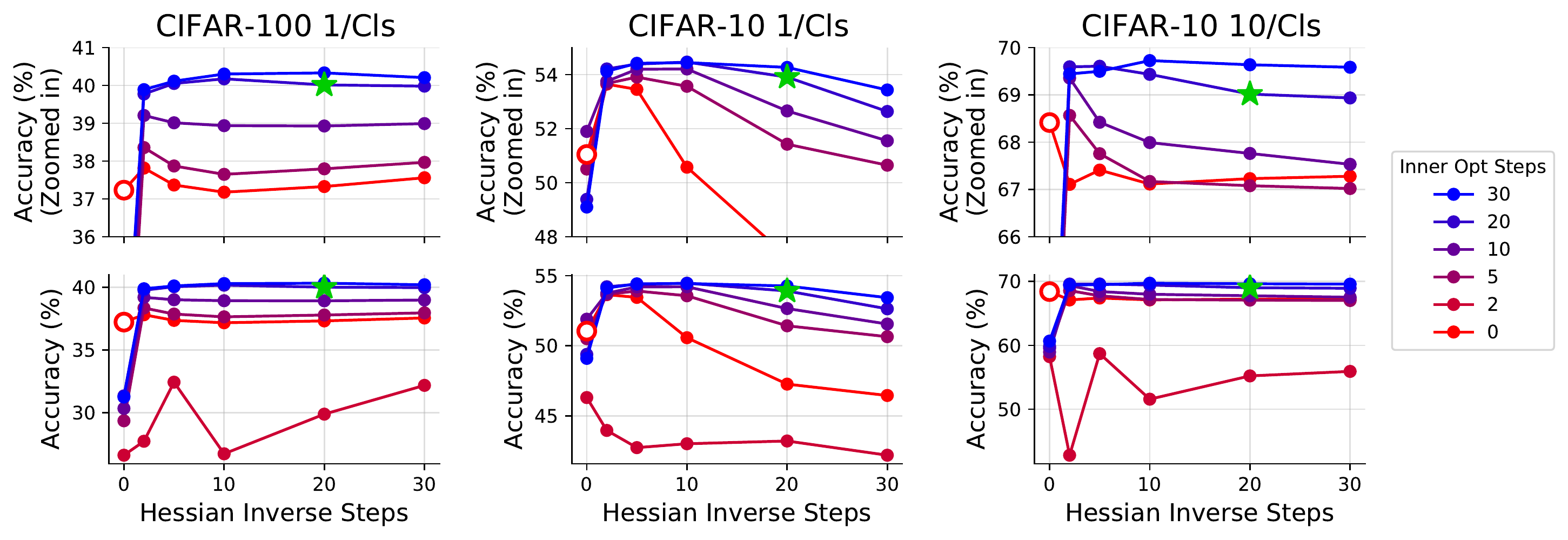}
\caption{The effect of $\nin$ and $\nhes$ on the distillation accuracy for CIFAR-100 1 Img/Cls, and CIFAR-10 with 1 and 10 Img/Cls. The green star denotes the hyperparameter configuration used throughout the paper ($\nin = 20, \nhes = 20$), while the red circle denotes using no implicit gradients and only direct gradients. The top row is the same as the bottom row except zoomed in. There is a clear advantage to using more inner optimization steps, provided that we account for the implicit gradient ($\nhes > 0$)}
\label{fig:cg_ablation_accuracy}
\end{center}
\vskip -0.2in
\end{figure*}

Specifically, for CIFAR-100 1 Img/Cls, CIFAR-10 1 Img/Cls, and CIFAR-10 10 Img/Cls, we rerun our algorithm with $\nin, \nhes \in \{0, 2, 5, 10, 20, 30\}$, and report the resulting runtime per iteration in \cref{fig:cg_ablation_time_per_iter} and accuracy in \cref{fig:cg_ablation_accuracy}, with the green stars, indicate our chosen hyperparameter configuration $\nin = \nhes = 20$, and red circle in \cref{fig:cg_ablation_accuracy} indicating the case where $\nin = \nhes = 0$. From \cref{fig:cg_ablation_time_per_iter}, we see that runtime is linear in both $\nin$ and $\nhes$, in line with our expectations. This has a large impact on the runtime of the algorithm, as for CIFAR-100 with 1 Img/Cls, the runtime per iteration is from 150ms to 760ms per iteration with the largest hyperparameter configurations.

In terms of accuracy, we see from \cref{fig:cg_ablation_accuracy}, \textbf{running inner optimization provides a clear performance benefit, only if we take into account the implicit gradient}. The $\nin = \nhes = 0$ (baseline) configuration achieves relatively high performance, and as discussed in \cref{sec:rcig_description}, this corresponds to the same algorithm as FRePo (provided that we use an MSE rather than Platt outer loss) and uses information from only the last layer. This method only leverages the information in the final layer weights and only has the direct gradient component in \cref{eq:full_meta_gradient}. If we set $\nin > 0$ but $\nhes = 0$, we see a clear performance drop compared to the baseline configuration, as it corresponds to ignoring the implicit gradient. However, adding a small number of Hessian Inverse steps (as few as 2) allows methods with $\nin > 0$, to exceed the base configuration's performance. When $\nin$ is small, larger $\nhes$ values hurt performance, as the implicit gradient formula only makes sense provided that we are at a minimum, which is not the case when $\nin$ is small. Finally, we observe that $\nin = 2$ sees very poor performance in general. 

\begin{figure}[ht]
\vskip 0.1in
\begin{center}
\includegraphics[width = \linewidth]{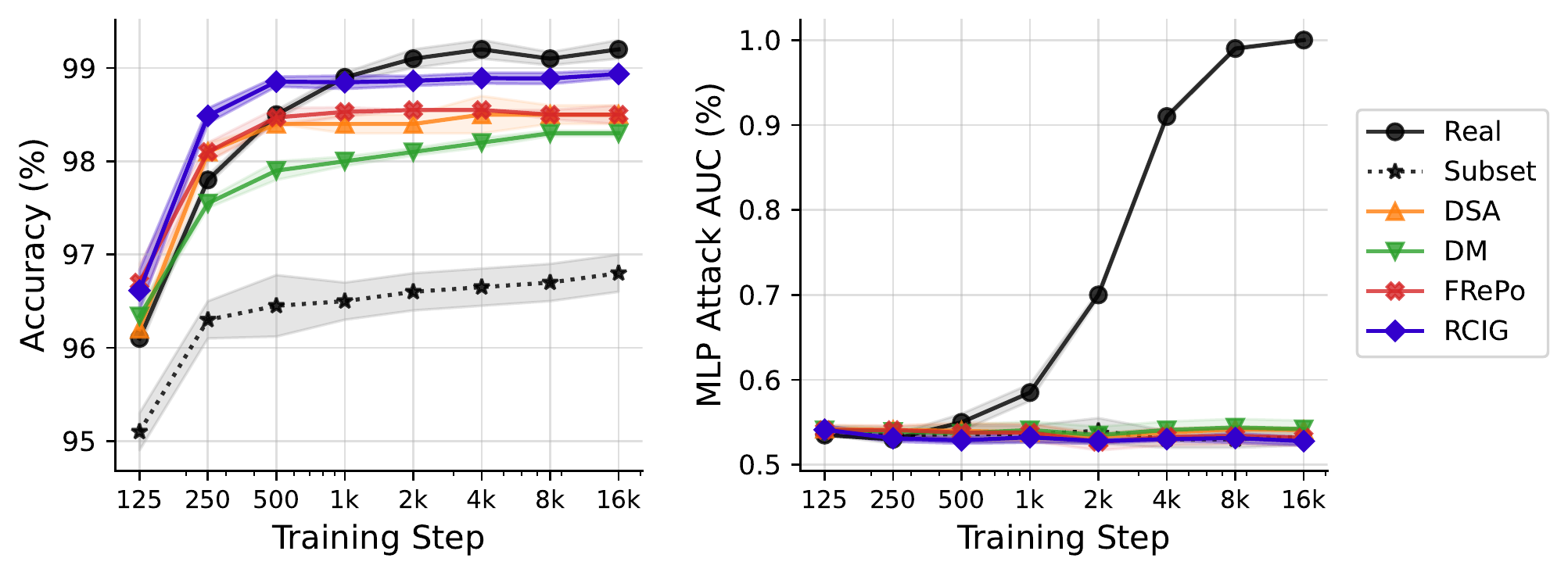}
\caption{Test accuracy and MLP Attack AUC for models trained on distilled data. Training on distilled data is not vulnerable to MIA attacks while training on real data leaks information 
}
\label{fig:mia_results}
\end{center}
\vskip -0.2in
\end{figure}

\textbf{4.5. Application: Privacy Preservation.} Membership inference attacks (MIA) \citep{MIA} try to determine whether a particular example was used to train a machine learning model. As training data can contain sensitive user information, it is critical that adversaries cannot infer what these training examples were. There exist many methods to defend against these attacks, such as adding gradient noise during training \citep{deep_learning_with_DP_OG_paper}, or strategic regularization \citep{dp_with_adversarial_reg}. Recently it has been shown that Dataset Distillation is a viable empirical defense strategy \citep{dd_with_dp}\footnote{see \citet{DD_with_DP_gets_rekt} for a more careful discussion}, as training neural networks on distilled data in practice resists membership inference attacks queried on the original training set.

Here we verify that RCIG also works as a viable defense strategy. We repeat the experimental procedure of \citet{frepo} and distill 10000 images of MNIST or Fashion-MNIST to 500 images. We run five (Threshold, Logistic-Regression (LR), Multi-Layered Perceptron (MLP), Random Forest (RF), and K-nearest-neighbor (KNN) attacks from Tensorflow-Privacy on neural networks trained on these 500 distilled images. We report the AUC of these attacks and resulting test accuracy in \cref{tab:mia_results_table} after training for 16k iterations. More experimental details are in Appendix. We see that RCIG achieves the highest accuracy of all the dataset distillation approaches, but remains resistant to the five attack strategies.

\section{Discussions, Limitations, and Conclusion}

In this paper, we presented RCIG, a dataset distillation algorithm based on implicit gradients, which achieved state-of-the-art performance on a wide variety of benchmarks. To derive our method, we first considered implicit gradients, then showed that linearizing the inner objective to make the problem convex improves performance, owing to the non-convexity of deep neural networks. Then we show that exploiting the neural network structure lets us more quickly converge to near-optimal solutions. We verified the efficiency of our approach on a wide variety of datasets, obtaining $\mathbf{15.6 \%}$ accuracy on ImageNet with 1 image per class, doubling the performance of prior art.

While our approach performs well, as discussed, our method can overfit on datasets with fewer training samples such as CUB-200. 
Future work could look at mitigating this overfitting issue. As we presented in \cref{sec:subsampling}, and showed in practice in \cref{sec:subsampling_ablation}, we can scale our algorithm to larger datasets without increasing the memory footprint by subsampling the backward pass. This method is still limited as it still requires a full forwards pass on the full distilled dataset. Future work can look at removing this limitation.

\section{Acknowledgements}
Research was sponsored by the United States Air Force Research Laboratory and the United States Air Force Artificial Intelligence Accelerator and was accomplished under Cooperative Agreement Number FA8750-19-2-1000. The views and conclusions contained in this document are those of the authors and should not be interpreted as representing the official policies, either expressed or implied, of the United States Air Force or the U.S. Government. The U.S. Government is authorized to reproduce and distribute reprints for Government purposes notwithstanding any copyright notation herein. The research was also funded in part by the AI2050 program at Schmidt Futures (Grant G-22-63172) and Capgemini SE.

\bibliography{main}
\bibliographystyle{icml2023}

\newpage
\appendix
\onecolumn

\section{Memory Requirements and Subsampling Ablation}
\label{sec:subsampling_ablation}

\begin{figure*}[h!]
\vskip 0.0in
\begin{center}
\subfloat[Full forward pass]{
\resizebox{!}{1.4in}{
\begin{tikzpicture}[node distance=2cm]
\node (x1) [graphnode]{$x_0$};
\node (x2) [graphnode, below =\nodevertical of x1]{$x_1$};
\node (x3) [graphnode, below =\nodevertical of x2]{$x_2$};

\node (fill_in) [emptynode, below =\nodevertical of x3]{$\vdots$};
\node (xs1) [graphnode, below =\nodevertical of fill_in]{$x_{|S| - 1}$};
\node (xs) [graphnode, below =\nodevertical of xs1]{$x_{|S|}$};

\node (h1) [hiddennode, right =\nodehorizontal of x1]{$h_1$};
\node (h2) [hiddennode, right =\nodehorizontal of x2]{$h_2$};
\node (h3) [hiddennode, right =\nodehorizontal of x3]{$h_3$};
\node (hs1) [hiddennode, right =\nodehorizontal of xs1]{$h_{|S| - 1}$};
\node (hs) [hiddennode, right =\nodehorizontal of xs]{$h_{|S|}$};

\node (loss) [endnode, right =\nodehorizontallong of $(x1.east)!0.5!(xs.east)$]{$\mathcal{L}$};

\draw [forwardarrow] (x1) -- node[anchor=south] {$\theta$} (h1);
\draw [forwardarrow] (x2) -- node[anchor=south] {$\theta$} (h2);
\draw [forwardarrow] (x3) -- node[anchor=south] {$\theta$} (h3);
\draw [forwardarrow] (xs1) -- node[anchor=south] {$\theta$} (hs1);
\draw [forwardarrow] (xs) -- node[anchor=south] {$\theta$} (hs);

\draw [forwardarrow] (h1) -- node[anchor=south] {} (loss);
\draw [forwardarrow] (h2) -- node[anchor=south] {} (loss);
\draw [forwardarrow] (h3) -- node[anchor=south] {} (loss);
\draw [forwardarrow] (hs1) -- node[anchor=south] {} (loss);
\draw [forwardarrow] (hs) -- node[anchor=south] {} (loss);

\draw [decorate,decoration = {calligraphic brace, mirror, amplitude = 10pt}, thick] ($(-1,0) + (x1.north)$) -- node[anchor=east, inner sep=18pt] {\fontsize{20pt}{20pt}\selectfont $|S|$} ($(-1,0) + (xs.south)$);
\end{tikzpicture}
}
}
\subfloat[Standard full backwards pass \\ to all images]{
\resizebox{!}{1.4in}{
\begin{tikzpicture}[node distance=2cm]
\node (x1) [graphnode]{$x_0$};
\node (x2) [graphnode, below =\nodevertical of x1]{$x_1$};
\node (x3) [graphnode, below =\nodevertical of x2]{$x_2$};

\node (fill_in) [emptynode, below =\nodevertical of x3]{$\vdots$};
\node (xs1) [graphnode, below =\nodevertical of fill_in]{$x_{|S| - 1}$};
\node (xs) [graphnode, below =\nodevertical of xs1]{$x_{|S|}$};

\node (h1) [hiddennode, right =\nodehorizontal of x1]{$h_1$};
\node (h2) [hiddennode, right =\nodehorizontal of x2]{$h_2$};
\node (h3) [hiddennode, right =\nodehorizontal of x3]{$h_3$};
\node (hs1) [hiddennode, right =\nodehorizontal of xs1]{$h_{|S| - 1}$};
\node (hs) [hiddennode, right =\nodehorizontal of xs]{$h_{|S|}$};

\node (loss) [endnode, right =\nodehorizontallong of $(x1.east)!0.5!(xs.east)$]{$\mathcal{L}$};

\begin{scope}[transform canvas={yshift=+0.6mm}]
\draw [forwardarrow] (x1) -- node[anchor=south] {$\theta$} (h1);
\draw [forwardarrow] (x2) -- node[anchor=south] {$\theta$} (h2);
\draw [forwardarrow] (x3) -- node[anchor=south] {$\theta$} (h3);
\draw [forwardarrow] (xs1) -- node[anchor=south] {$\theta$} (hs1);
\draw [forwardarrow] (xs) -- node[anchor=south] {$\theta$} (hs);
\end{scope}

\begin{scope}[transform canvas={yshift=+0.6mm}]
\draw [forwardarrow] (h1) -- node[anchor=south] {} (loss);
\draw [forwardarrow] (h2) -- node[anchor=south] {} (loss);
\draw [forwardarrow] (h3) -- node[anchor=south] {} (loss);
\draw [forwardarrow] (hs1) -- node[anchor=south] {} (loss);
\draw [forwardarrow] (hs) -- node[anchor=south] {} (loss);
\end{scope}

\begin{scope}[transform canvas={yshift=-0.6mm}]
\draw [backwardarrow] (h1) -- node[anchor=south] {} (x1);
\draw [backwardarrow] (h2) -- node[anchor=south] {} (x2);
\draw [backwardarrow] (h3) -- node[anchor=south] {} (x3);
\draw [backwardarrow] (hs1) -- node[anchor=south] {} (xs1);
\draw [backwardarrow] (hs) -- node[anchor=south] {} (xs);
\end{scope}

\begin{scope}[transform canvas={yshift=-0.6mm}]
  \draw [backwardarrow] (loss) -- node[anchor=south] {} (h1);
\draw [backwardarrow] (loss) -- node[anchor=south] {} (h2);
\draw [backwardarrow] (loss) -- node[anchor=south] {} (h3);
\draw [backwardarrow] (loss) -- node[anchor=south] {} (hs1);
\draw [backwardarrow] (loss) -- node[anchor=south] {} (hs);
\end{scope}

\draw [decorate,decoration = {calligraphic brace, mirror, amplitude = 10pt}, thick] ($(-1,0) + (x1.north)$) -- node[anchor=east, inner sep=18pt] {\fontsize{20pt}{20pt}\selectfont $|S|$} ($(-1,0) + (xs.south)$);
\end{tikzpicture}
}
}
\subfloat[Unbiased backwards pass \\ to only subset of images]{
\resizebox{!}{1.4in}{
\begin{tikzpicture}[node distance=2cm]
\node (x1) [graphnode, opacity = \fadeopacity]{$x_0$};
\node (x2) [graphnode, below =\nodevertical of x1]{$x_1$};
\node (x3) [graphnode, below =\nodevertical of x2, opacity = \fadeopacity]{$x_2$};

\node (fill_in) [emptynode, below =\nodevertical of x3, opacity = \fadeopacity]{$\vdots$};
\node (xs1) [graphnode, below =\nodevertical of fill_in, opacity = \fadeopacity]{$x_{|S| - 1}$};
\node (xs) [graphnode, below =\nodevertical of xs1]{$x_{|S|}$};

\node (h1) [hiddennode, right =\nodehorizontal of x1, opacity = \fadeopacity]{$h_1$};
\node (h2) [hiddennode, right =\nodehorizontal of x2]{$h_2$};
\node (h3) [hiddennode, right =\nodehorizontal of x3, opacity = \fadeopacity]{$h_3$};
\node (hs1) [hiddennode, right =\nodehorizontal of xs1, opacity = \fadeopacity]{$h_{|S| - 1}$};
\node (hs) [hiddennode, right =\nodehorizontal of xs]{$h_{|S|}$};

\node (loss) [endnode, right =\nodehorizontallong of $(x1.east)!0.5!(xs.east)$]{$\mathcal{L}$};

\begin{scope}[transform canvas={yshift=+0.6mm}]
\draw [forwardarrow_faded] (x1) -- node[anchor=south] {$\theta$} (h1);
\draw [forwardarrow] (x2) -- node[anchor=south] {$\theta$} (h2);
\draw [forwardarrow_faded] (x3) -- node[anchor=south] {$\theta$} (h3);
\draw [forwardarrow_faded] (xs1) -- node[anchor=south] {$\theta$} (hs1);
\draw [forwardarrow] (xs) -- node[anchor=south] {$\theta$} (hs);
\end{scope}

\begin{scope}[transform canvas={yshift=+0.6mm}]
\draw [forwardarrow_faded] (h1) -- node[anchor=south] {} (loss);
\draw [forwardarrow] (h2) -- node[anchor=south] {} (loss);
\draw [forwardarrow_faded] (h3) -- node[anchor=south] {} (loss);
\draw [forwardarrow_faded] (hs1) -- node[anchor=south] {} (loss);
\draw [forwardarrow] (hs) -- node[anchor=south] {} (loss);
\end{scope}

\begin{scope}[transform canvas={yshift=-0.6mm}]
\draw [backwardarrow] (h2) -- node[anchor=south] {} (x2);
\draw [backwardarrow] (hs) -- node[anchor=south] {} (xs);
\end{scope}

\begin{scope}[transform canvas={yshift=-0.6mm}]
\draw [backwardarrow] (loss) -- node[anchor=south] {} (h2);
\draw [backwardarrow] (loss) -- node[anchor=south] {} (hs);
\end{scope}
\draw [decorate,decoration = {calligraphic brace, mirror, amplitude = 10pt}, thick] ($(-1,0) + (x2.north)$) -- node[anchor=east, inner sep=18pt] {\fontsize{20pt}{20pt}\selectfont $\nsub$} ($(-1,0) + (xs.south)$);
\end{tikzpicture}
}
}

\caption{Our proposed subsampling scheme. First, we perform a full forwards pass using all distilled images in (a). Typically, one performs a backward pass to all distilled images (b), but this requires retaining the entire computation graph which is expensive for large coresets. Instead, we backpropagate only through to $\nsub < |S|$ distilled images (c), using values computed using the full forward pass. Due to the exchangeability of the coreset images, this results in unbiased gradients.}
\label{fig:subsampling}
\end{center}
\vskip -0.2in
\end{figure*}
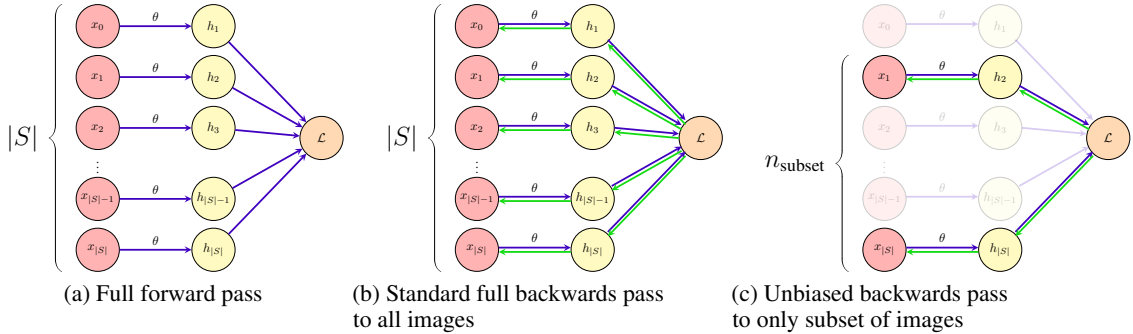

\begin{figure}[h]
\vskip 0.1in
\begin{center}
\includegraphics[width = 0.7 \linewidth]{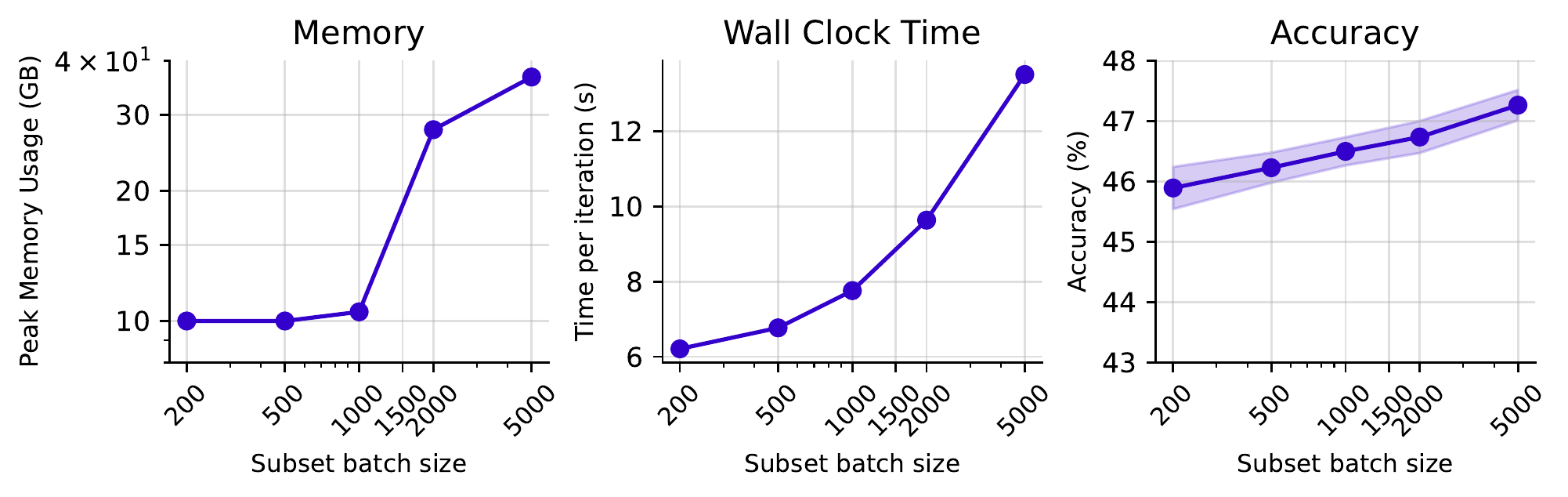}
\caption{The effect of $n_\textrm{subset}$ on memory consumption, wall clock time, and accuracy on CIFAR-100 with 50 Img/Cls. Memory consumption increases significantly with larger $n_\textrm{subset}$ and time per iteration is approximately linear in the $\nsub$. In contrast, $\nsub$, has minimal effect on distillation accuracy.}
\label{fig:cf100_ablation}
\end{center}
\vskip -0.2in
\end{figure}

In \cref{sec:subsampling}, we claimed that we can obtain unbiased gradient estimates by performing a full forward pass on the entire distilled dataset, and only performing a backward pass on a subset. Here we verify this claim by varying the subset size, $n_\textrm{subset}$ for the backward pass. We test this for CIFAR-100 with 50 images per class, which results in the largest number of distilled images at 5000 of all the experiments in \cref{sec:standard_results}. In \cref{sec:standard_results}, we set $n_\textrm{subset}$ to be 2000, 40\% of the full dataset, but here we let $n_\textrm{subset} \in \{200, 500, 1000, 2000, 5000\}$, and measure the resulting memory consumption, wall clock time, and resulting accuracy in \cref{fig:cf100_ablation}, distilling for 4000 iterations. We see that larger $\nsub$ consumes significantly more memory, and that wall clock time is linear in $\nsub$. Despite this, accuracy is only slightly affected, moving from $45.9 \pm 0.4 \%$ to $47.3 \pm 0.3$ for $\nsub = 200$, and $\nsub = 5000$, respectively, despite $\nsub = 5000$ requiring 26.7Gb compared to $\nsub = 200$'s 10.0Gb. The small performance drop could be due to increased gradient variance associated with the estimator, or that the effective number of dataset updates is fewer for small $\nsub$, as only a subset of distilled images is updated per iteration.

\section{Privacy Preservation: Membership Inference Attacks. More details.}
\begin{figure}[h]
\vskip 0.1in
\begin{center}
\includegraphics[width = 0.7\linewidth]{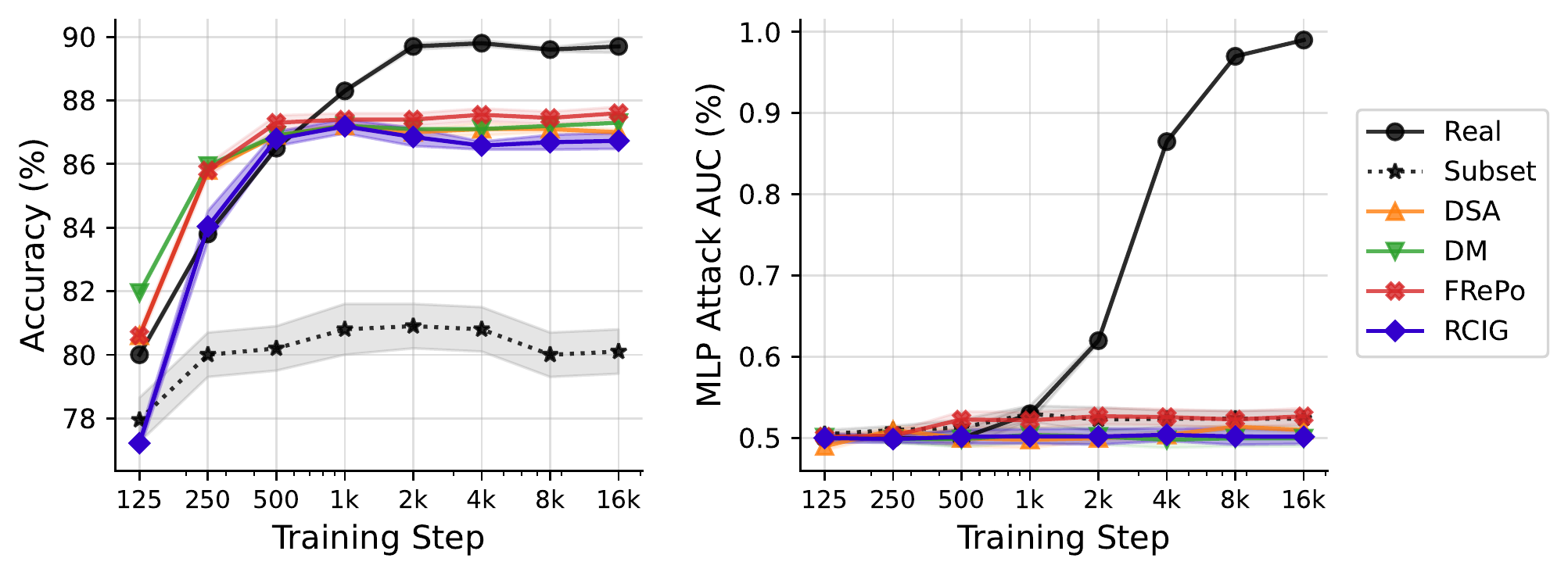}
\caption{Test accuracy and MLP Attack AUC for models trained on distilled data for Fashion-MNIST. Training on distilled data is not vulnerable to MIA attacks while training on real data leaks information 
}
\label{app:fig:fashion_mia}
\end{center}
\vskip -0.2in
\end{figure}

For the non-member examples we use the test set of 10k images so that an AUC of 0.5 effectively means that the attack is a random guess. Additionally, we plot the MLP attack AUC and accuracy as a function of training iteration. We report baselines of using the original training set (real), a random subset of 500 images, and the DSA, DM, FRePo dataset distillation algorithms. We also report results on Fashion-MNIST in \cref{app:fig:fashion_mia} and \cref{app:tab:fashion_mia}. Results for baseline algortihms are taken from \citet{frepo}. We also show the resulting curves for CIFAR-10 and CIFAR-100 for RCIG compared to the real dataset baselines. In this two more complex datasets, there is a better advantage to training with distilled data at a given privacy budget.

\begin{table*}[t]
\centering
\caption{AUC of five MIA attack strategies for neural networks trained on distilled data on Fashion-MNIST. 
(n=25)}
\begin{adjustbox}{width=0.7\textwidth}
\begin{tabular}{lcccccc} \toprule
              & \multirow{2}{*}{Test Acc (\%)} & \multicolumn{5}{c}{Attack AUC}  \\ \cmidrule{3-7}
              &                                & Threshold & LR & MLP & RF & KNN \\ \midrule
Real	&	$89.7 \pm 0.2 $	&	$0.99 \pm 0.01 $	&	$0.99 \pm 0.00 $	&	$0.99 \pm 0.00 $	&	$0.99 \pm 0.00 $	&	$0.98 \pm 0.00$	\\
Subset	&	$81.1 \pm 0.7 $	&	$0.53 \pm 0.01 $	&	$0.51 \pm 0.01$	&	$0.52 \pm 0.01 $	&	$0.52 \pm 0.01 $	&	$0.53 \pm 0.00$	\\
DSA	&	$87.0 \pm 0.1 $	&	$0.51 \pm 0.00 $	&	$0.51 \pm 0.01 $	&	$0.51 \pm 0.01 $	&	$0.52 \pm 0.01 $	&	$0.51 \pm 0.01$	\\
DM	&	$87.3 \pm 0.1 $	&	$0.52 \pm 0.00 $	&	$0.51 \pm 0.01 $	&	$0.50 \pm 0.01 $	&	$0.52 \pm 0.01 $	&	$0.51 \pm 0.01$	\\
FRePo	&	$87.6 \pm 0.2 $	&	$0.52 \pm 0.00 $	&	$0.53 \pm 0.01 $	&	$0.53 \pm 0.01 $	&	$0.53 \pm 0.01 $	&	$0.52 \pm 0.00$	\\
RCIG	&	$86.7 \pm 0.3$	&	$0.50 \pm 0.01$	&	$0.50 \pm 0.01$	&	$0.50 \pm 0.01$	&	$0.50 \pm 0.01$	&	$0.50 \pm 0.00$	\\ \bottomrule
\end{tabular}
\end{adjustbox}
\label{app:tab:fashion_mia}
\end{table*}

\begin{figure}[h]
\vskip 0.1in
\begin{center}
\includegraphics[width = 0.49\linewidth]{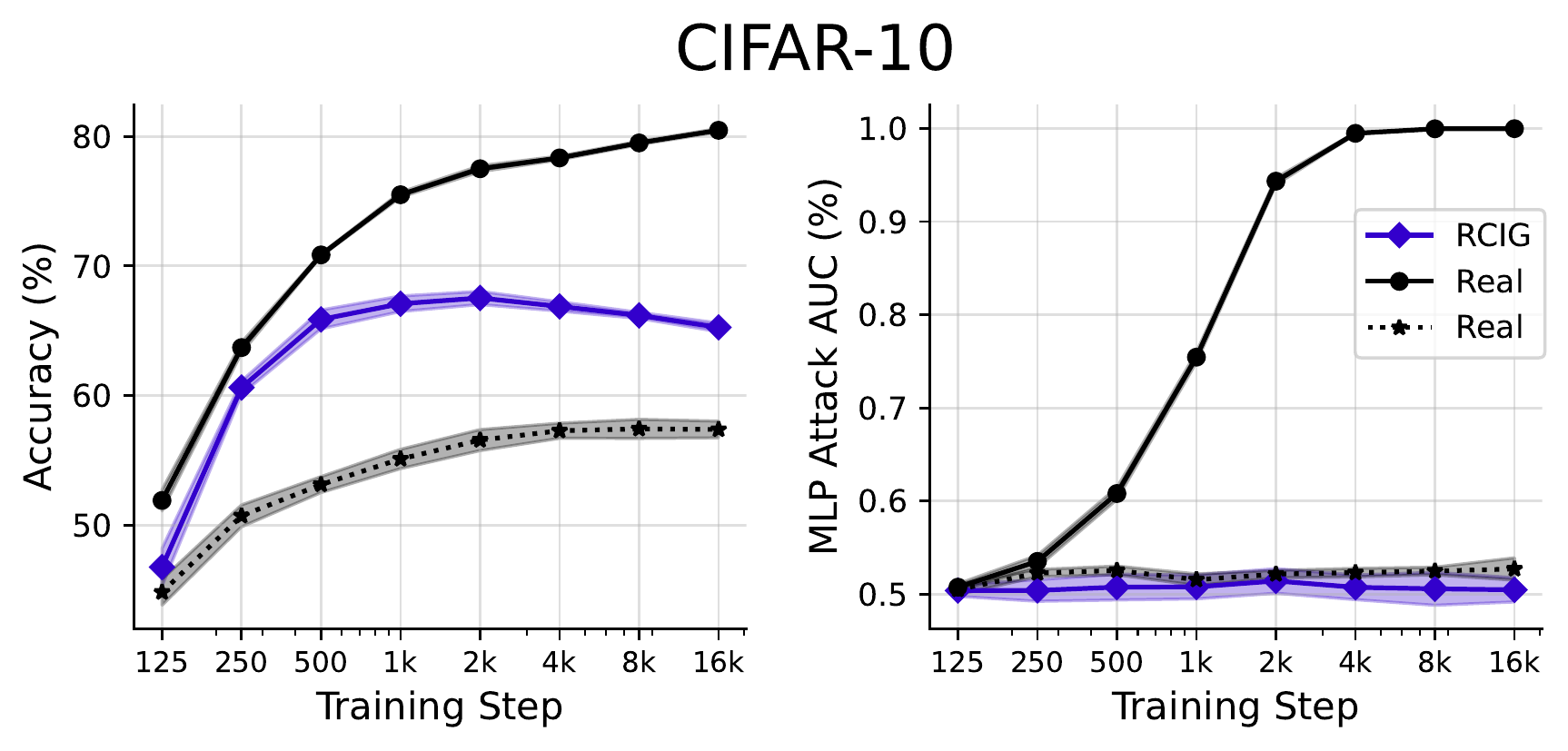}
\includegraphics[width = 0.49\linewidth]{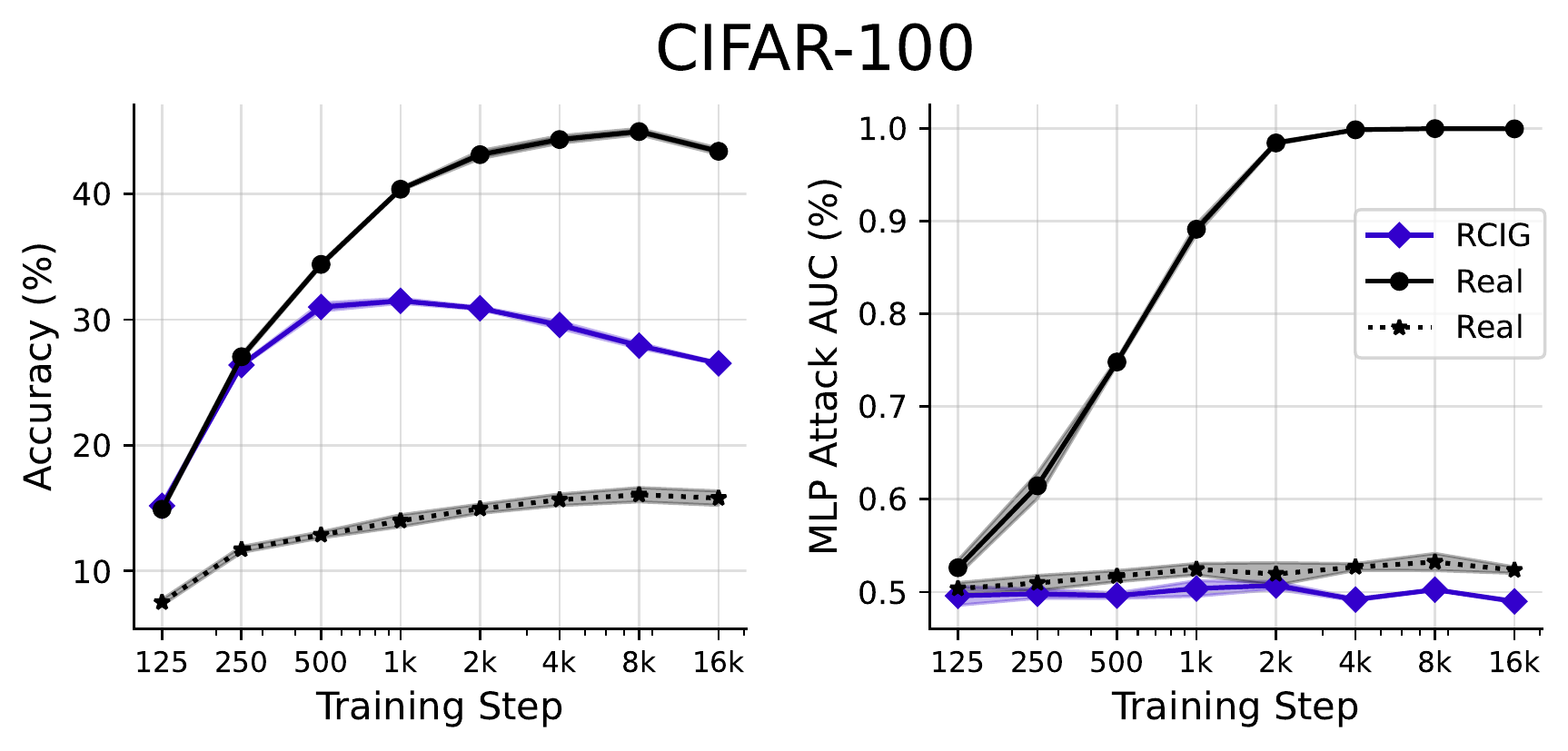}
\caption{Test accuracy and MLP Attack AUC for models trained on distilled data for CIFAR-10 (left) and CIFAR-100 (right).
}
\label{app:fig:cifar10_mia}
\end{center}
\vskip -0.2in
\end{figure}

\section{Implementation details}
\textbf{Libraries and Hardware.} 
Code is implemented using the libraries JAX, Optax, and Flax \citep{jax2018github, deepmind2020jax, flax}. We use a mix of Nvidia Titan RTXs with 24Gb, RTX 4090s with 24Gb, and Quadro RTX A6000s with 48Gb VRAM. The training time per iteration plots in \cref{fig:cg_ablation_time_per_iter} and \cref{fig:cf100_ablation} are run on an RTX 4090.

\textbf{Optimizers, Hyperparameter selection.} 
For the $\lambda$ $L_2$ regularization term, for depth 3 models we used $\lambda = 0.0005 \times |S|$, for depth 4 we use $\lambda = 0.005 \times |S|$ and depth 5 we use $\lambda = 0.05 \times |S|$. We did not find that this had a major impact in terms of performance, but Hessian-Inverse computation could be unstable if it is too low.

For the coreset optimizer, we use Adabelief \citep{adabelief} optimizer with learning rate 0.003 for the coreset images and labels, and a learning rate of 0.03 for $\log \tau$, the Platt scaling loss temperature. For inner optimization and Hessian inverse computation, we use Adam optimizer \citep{adam}, with learning rates $\ain$ and $\ahes$, we perform a small tuning round before beginning coreset optimization. We increase both learning rates as high as possible such that the optimization of loss of the inner problem and Hessian ivnerse loss (\cref{eq:hinv_loss}) are monotonically decreasing (this is done automatically). During the course of training if we observe that either loss diverges, we reduce the corresponding learning rate by 10\%. In general these learning rates should be as high as possible such that optimization is stable.

\textbf{Dataset Preprocessing.} 
We use the same ZCA regularization with regularization strength $\lambda = 0.1$ as \citet{frepo} for RGB datasets and standard preprocessing for grayscale datasets.

\textbf{Subsampling.} 
As discussed in \cref{sec:subsampling} and \cref{sec:subsampling_ablation}, we can run out of memory for very large datasets. We use subsampling with batch sizes of 2000 for CIFAR-100 with 50 img/cls, and 500 for Tiny-ImageNet 10 img/cls and resized ImageNet. In general this should be as large as possible that fits in the memory.

\textbf{Evaluation.} 
During evaluation, we train neural networks for 1000 iterations if $|S| = 10$, otherwise for 2000 iterations. We used Adam optimizer with a learning rate of 0.0001. In line with prior work \citep{frepo}, we use a learning rate schedule with 500 iterations of linear warm up following be a cosine decay. During testing evaluation we either use no data augmentation, or we use flip, color, crop, rotate, translate and cutout data augmentation for RGB datasets, or crop, rotate, translate, cutout for grayscale ones. This is the same procedure used in prior work.

\textbf{Models.}
Unless otherwise stated, we used the same Convolutional network used in \citet{frepo}, which doubles the number of convolutional filters for every layer. For datasets of resolution $32 \times 32$ or $28\times 28$, we use depth 3, for datasets of resolution $64\times 64$ (Tiny-ImageNet, Resized ImageNet), we use depth 4, for $128\times 128$ we use depth 5, in line with prior work. We do not use normalization during training, unless otherwise stated. For models of depth 3, during training with replaced with ReLUs with softplus activations with temperature 60, this did not make a major difference in the final result but sped up convergence, as it allows the gradients to be smoother as opposed to a step function. During evaluation we used standard ReLUs.

\textbf{Initialization}
We initialize distilled datasets with samples from the original dataset. We note that for the high resolution datasets, the distilled images still resemble their initializations (see \cref{app:vis}). For low resolution datasets, we typically found that random noise initialization slowed convergence speed, but did not significantly affect the final result. However, for high resolution datasets, performance is harmed with noise initialization. For example, for ImageWoof on 10 ipc, we found that performance drops from $42.2\pm 0.7\%$ with real image initialization to $38.5\pm0.6\%$ with random noise initialization. Better initialization strategies could be the subject of future work.

\textbf{Miscellaneous.} 
We use the same flipping strategy employed in \citet{frepo} to avoid the mirroring effect for RGB datasets. We use this whenever $|S| \leq 1000$. For MNIST 50 img/cls and Fashion-MNIST 50 img/cls, we use 64-bit computation when computing the matrix inverse in \cref{eq:optimal_theta_f}, as we found otherwise it would be unstable. Note that the network forward pass and everything else is still 32-bit. Every other dataset uses the default 32-bit. For \cref{tab:linearization_ablation,tab:main_res_table,tab:imagenet_acc_table,tab:arch_transfer} we repeat distillation for 3 independent coreset initializations, and evaluate each on five networks (15 total evaluations). For MIA results, we perform distillation on 5 independents chunks of the dataset, and evaluate on 5 models for each (25 total evaluations).

\subsection{FRePo Code Error}
\label{app:frepo_bug}
We noted in \cref{tab:arch_transfer} that we achieved different results on FRePo \citep{frepo} using their public code repository. Additionally we found a bug where in line 202-203 in \url{https://github.com/yongchao97/FRePo/blob/master/script/eval.py}, where their code sets all evaluation normalization layers to no normalization. The corrected results are in \cref{tab:arch_transfer}.

\section{Additional Results}
\subsection{Training Curves}

\begin{figure}[h]
\vskip 0.1in
\begin{center}
\includegraphics[width = 0.9\linewidth]{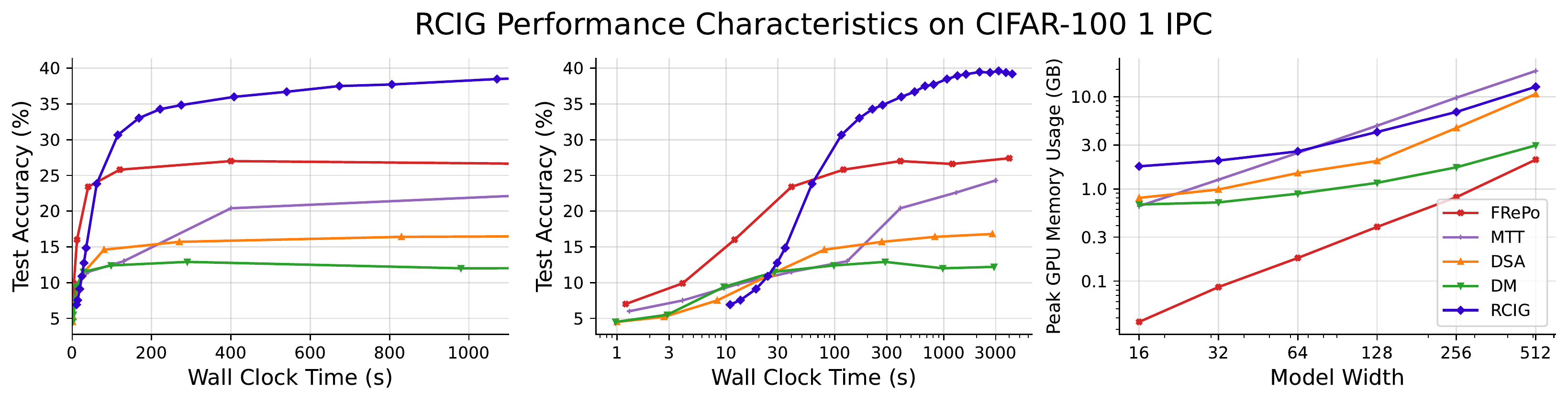}
\caption{Test accuracy vs. wall clock time for CIFAR-100 distillation with 1 IPC, with linear (left) and log (center) axes, and the memory consumption of RCIG (right) at different model widths. We use $w=128$ for all other experiments in the paper.}
\label{app:fig:training_curve}
\end{center}
\vskip -0.2in
\end{figure}

In \cref{app:fig:training_curve}, we report the test accuracy as a function of distillation wall clock time for RCIG and other algorithms at distilling CIFAR-100 with 1 IPC. RCIG quickly exceeds the performance of other algorithms within 100s. This experiment was run on a RTX 4090. In order to account for the stronger GPU and make our times comparable to those in \citet{frepo} we increase all measured times by $40\%$. We also show the memory requirements of RCIG in \cref{app:fig:training_curve}. RCIG typically requires more memory than other methods due to linearization, but memory requirements do not scale with the number of inner optimization steps or unroll steps.

\subsection{Total Training Time}

\begin{table*}[h]
\centering
 \begin{adjustbox}{width=0.8\textwidth}
\begin{tabular}{ccccc} \toprule
                            & Img/Cls & Total Training Time (h) & Time per iter (s) & Training iterations \\ \midrule
\multirow{3}{*}{MNIST} & 1	&	$0.67 \pm 0.01$	&	$0.16 \pm 0.00$	&	15000\\
                     & 10	&	$1.25 \pm 0.03$	&	$0.45 \pm 0.01$	&	10000\\
                     & 50	&	$2.75 \pm 0.00$	&	$1.98 \pm 0.00$	&	5000\\ \midrule
\multirow{3}{*}{F-MNIST} & 1	&	$0.67 \pm 0.01$	&	$0.16 \pm 0.00$	&	15000\\
                     & 10	&	$1.25 \pm 0.03$	&	$0.45 \pm 0.01$	&	10000\\
                     & 50	&	$2.67 \pm 0.06$	&	$1.92 \pm 0.04$	&	5000\\ \midrule
\multirow{3}{*}{CIFAR-10} & 1	&	$0.74 \pm 0.00$	&	$0.18 \pm 0.00$	&	15000\\
                     & 10	&	$1.82 \pm 0.07$	&	$0.65 \pm 0.02$	&	10000\\
                     & 50	&	$4.09 \pm 0.20$	&	$2.94 \pm 0.14$	&	5000\\ \midrule
\multirow{3}{*}{CIFAR-100} & 1	&	$1.88 \pm 0.07$	&	$0.68 \pm 0.03$	&	10000\\
                     & 10	&	$8.55 \pm 0.08$	&	$6.15 \pm 0.06$	&	5000\\
                     & 50	&	$11.21 \pm 0.65$	&	$10.09 \pm 0.58$	&	4000\\ \midrule
\multirow{2}{*}{T-ImageNet} & 1	&	$7.78 \pm 0.00$	&	$5.60 \pm 0.00$	&	5000\\
                     & 10	&	$6.43 \pm 0.46$	&	$11.58 \pm 0.82$	&	2000\\ \midrule
\multirow{2}{*}{CUB-200} & 1	&	$2.75 \pm 0.69$	&	$0.99 \pm 0.25$	&	10000\\
                     & 10	&	$9.58 \pm 3.22$	&	$17.24 \pm 5.80$	&	2000\\ \midrule
\multirow{2}{*}{ImageNette} & 1	&	$5.05 \pm 0.23$	&	$3.03 \pm 0.14$	&	6000\\
                     & 10	&	$12.62 \pm 0.03$	&	$15.15 \pm 0.04$	&	3000\\ \midrule
\multirow{2}{*}{ImageWoof} & 1	&	$4.78 \pm 0.00$	&	$2.87 \pm 0.00$	&	6000\\
                     & 10	&	$11.02 \pm 0.02$	&	$13.22 \pm 0.02$	&	3000\\ \midrule
\multirow{2}{*}{ImageNet} & 1	&	$4.75 \pm 0.34$	&	$8.55 \pm 0.61$	&	2000\\
                     & 2	&	$6.87 \pm 0.38$	&	$12.37 \pm 0.69$	&	2000\\ \bottomrule
\end{tabular}
\end{adjustbox}
\caption{Total runtime of RCIG (n=3)}
\label{app:tab:time_taken}
\end{table*}

In \cref{app:tab:time_taken} we report the total training time for or algorithm on the benchmarks. These experiments were run on Quadro RTX A6000s with 48Gb memory. This memory requirement is not necessary as we can reduce the memory consumption using the techniques described in \cref{sec:subsampling} and \cref{sec:subsampling_ablation}. There is high variance associated with the training times because some of the GPUs were run with reduced power limits.

\subsection{Evaluation with/without data augmentation}
In \cref{app:tab:da_acc} we report the evaluation accuracy of our algorithm when applying no data augment or data augmentation during test time. We observe that for large distilled datasets ($|S| > 200$), data augmentation seems to help, but hinders performance for smaller ones.

\begin{table*}[h]
\centering
 \begin{adjustbox}{width=0.8\textwidth}
\begin{tabular}{cccc} \toprule
                            & Img/Cls & No Data Augmentation & With Data Augmentation \\ \midrule
\multirow{3}{*}{MNIST} & 1	&	$\mathbf{ 94.7 \pm 0.5 }$	&	$ 92.7 \pm 0.8 $	\\
                     & 10	&	$\mathbf{ 98.9 \pm 0.0 }$	&	$ 98.7 \pm 0.1 $	\\
                     & 50	&	$ 99.1 \pm 0.1 $	&	$\mathbf{ 99.2 \pm 0.0 }$	\\ \midrule
\multirow{3}{*}{F-MNIST} & 1	&	$\mathbf{ 79.8 \pm 1.1 }$	&	$ 70.5 \pm 3.5 $	\\
                     & 10	&	$\mathbf{ 88.5 \pm 0.2 }$	&	$ 86.7 \pm 0.2 $	\\
                     & 50	&	$\mathbf{ 90.2 \pm 0.2 }$	&	$ 88.7 \pm 0.2 $	\\ \midrule
\multirow{3}{*}{CIFAR-10} & 1	&	$\mathbf{ 53.9 \pm 1.0 }$	&	$ 50.9 \pm 1.5 $	\\
                     & 10	&	$\mathbf{ 69.1 \pm 0.4 }$	&	$ 66.4 \pm 1.0 $	\\
                     & 50	&	$\mathbf{ 73.5 \pm 0.3 }$	&	$ 72.6 \pm 0.4 $	\\ \midrule
\multirow{3}{*}{CIFAR-100} & 1	&	$\mathbf{ 39.3 \pm 0.4 }$	&	$ 36.7 \pm 0.3 $	\\
                     & 10	&	$\mathbf{ 44.1 \pm 0.4 }$	&	$ 43.5 \pm 0.3 $	\\
                     & 50	&	$ 45.5 \pm 0.4 $	&	$\mathbf{ 46.7 \pm 0.3 }$	\\ \midrule
\multirow{2}{*}{T-ImageNet} & 1	&	$\mathbf{ 25.6 \pm 0.3 }$	&	$ 24.2 \pm 0.2 $	\\
                     & 10	&	$ 27.4 \pm 0.3 $	&	$\mathbf{ 29.4 \pm 0.2 }$	\\ \midrule
\multirow{2}{*}{CUB-200} & 1	&	$ 11.2 \pm 0.4 $	&	$\mathbf{ 12.1 \pm 0.2 }$	\\
                     & 10	&	$ 14.3 \pm 0.3 $	&	$\mathbf{ 15.7 \pm 0.3 }$	\\ \midrule
\multirow{2}{*}{ImageNette} & 1	&	$\mathbf{ 53.0 \pm 0.9 }$	&	$ 47.2 \pm 1.2 $	\\
                     & 10	&	$\mathbf{ 65.0 \pm 0.7 }$	&	$ 63.9 \pm 0.8 $	\\ \midrule
\multirow{2}{*}{ImageWoof} & 1	&	$\mathbf{ 33.9 \pm 0.6 }$	&	$ 27.0 \pm 2.1 $	\\
                     & 10	&	$\mathbf{ 42.2 \pm 0.7 }$	&	$ 40.2 \pm 0.6 $	\\ \midrule
\multirow{2}{*}{ImageNet} & 1	&	$ 14.6 \pm 0.2 $	&	$\mathbf{ 15.6 \pm 0.2 }$	\\
                     & 2	&	$ 15.9 \pm 0.1 $	&	$\mathbf{ 16.6 \pm 0.1 }$	\\ \bottomrule
\end{tabular}
\end{adjustbox}
\caption{Evaluation result of RCIG with or without using data augmentation. For large distilled datasets ($|S| > 200$), data augmentation seems to help, but hinders performance for smaller ones.}
\label{app:tab:da_acc}
\end{table*}

\section{Distilled Dataset Visualization}
\label{app:vis}
Here we show the resulting distilled dataset made by RCIG.

\begin{figure*}[h]
\vskip 0.1in
\begin{center}
\includegraphics[width = 0.7 \linewidth]{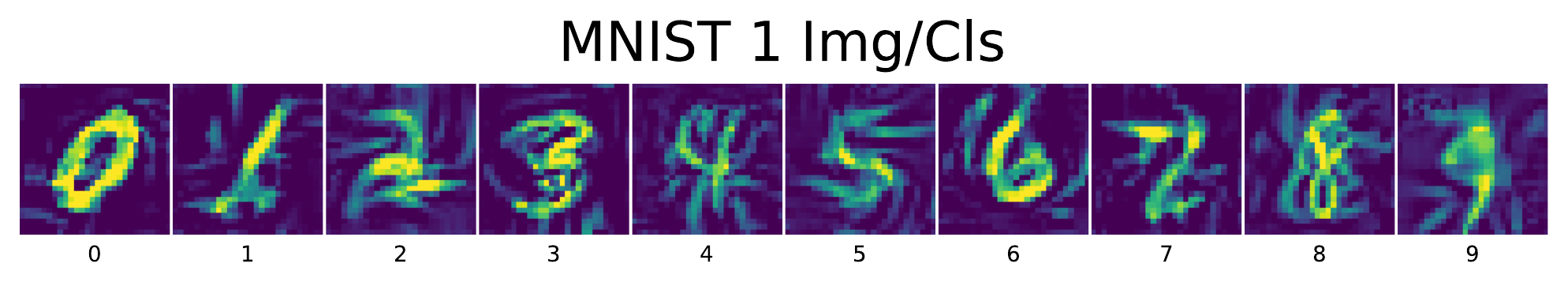}
\caption{RCIG distilled dataset on MNIST with 1 Img/Cls}
\end{center}
\vskip -0.2in
\end{figure*}

\begin{figure*}[h]
\vskip 0.1in
\begin{center}
\includegraphics[width = 0.7 \linewidth]{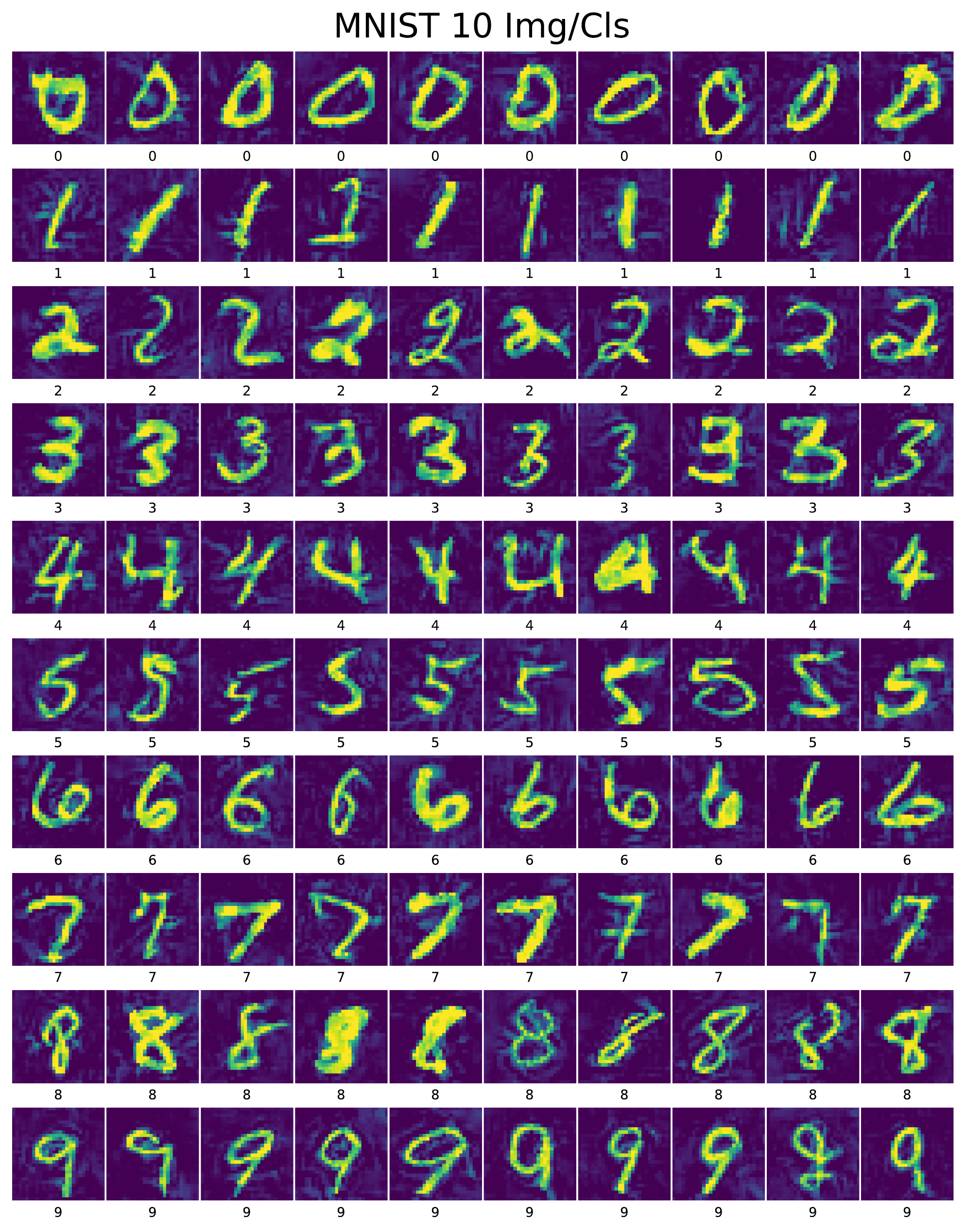}
\caption{RCIG distilled dataset on MNIST with 10 Img/Cls}
\end{center}
\vskip -0.2in
\end{figure*}

\begin{figure*}[h]
\vskip 0.1in
\begin{center}
\includegraphics[width = 0.7 \linewidth]{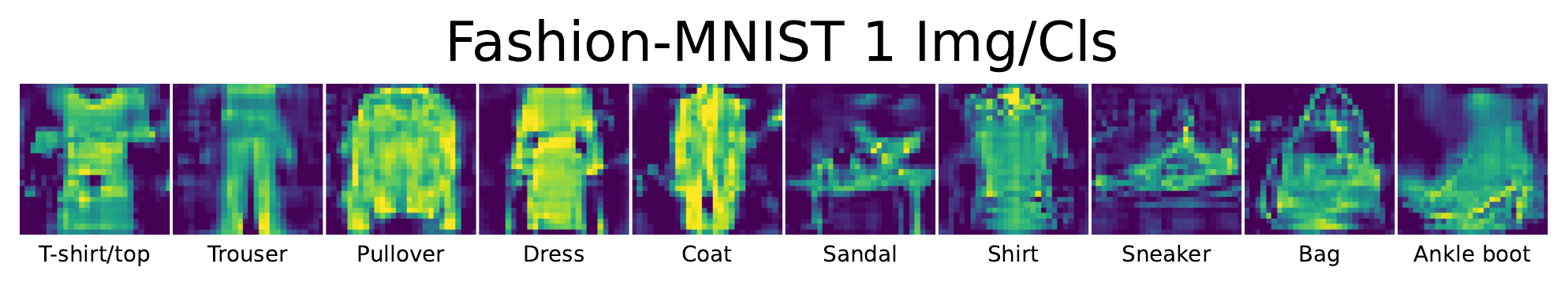}
\caption{RCIG distilled dataset on Fashion-MNIST with 1 Img/Cls}
\end{center}
\vskip -0.2in
\end{figure*}

\begin{figure*}[h]
\vskip 0.1in
\begin{center}
\includegraphics[width = 0.7 \linewidth]{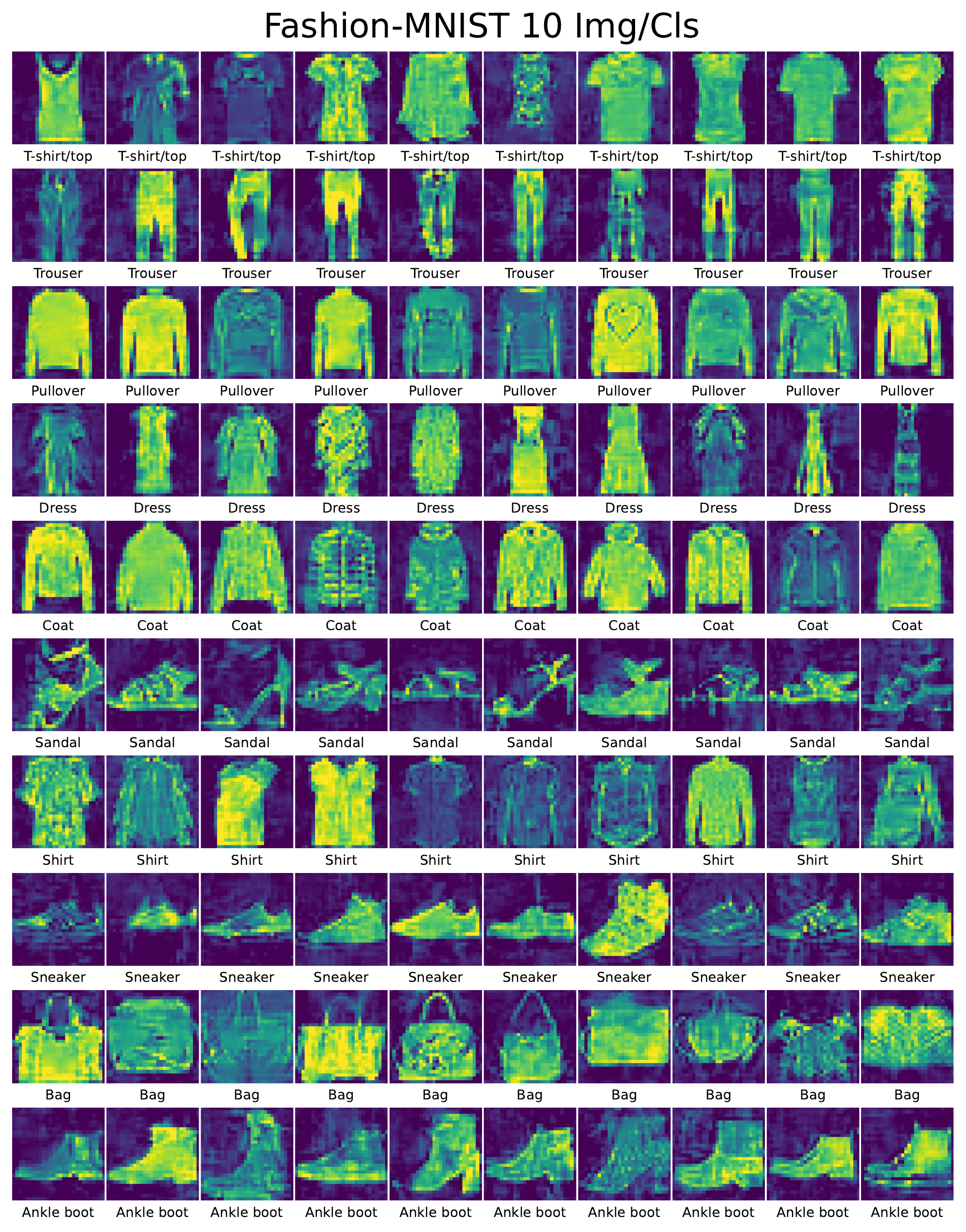}
\caption{RCIG distilled dataset on Fashion-MNIST with 10 Img/Cls}
\end{center}
\vskip -0.2in
\end{figure*}

\begin{figure*}[h]
\vskip 0.1in
\begin{center}
\includegraphics[width = 0.7 \linewidth]{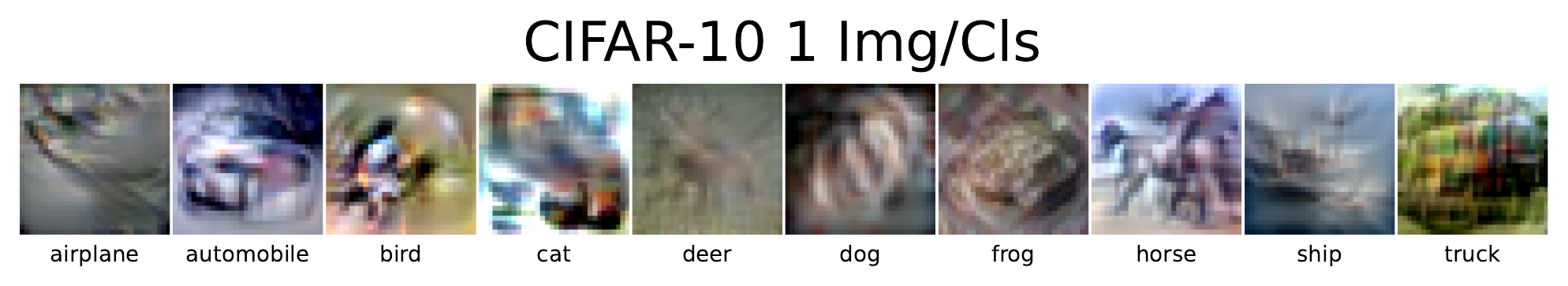}
\caption{RCIG distilled dataset on CIFAR-10 with 10 Img/Cls}
\end{center}
\vskip -0.2in
\end{figure*}

\begin{figure*}[h]
\vskip 0.1in
\begin{center}
\includegraphics[width = 0.7 \linewidth]{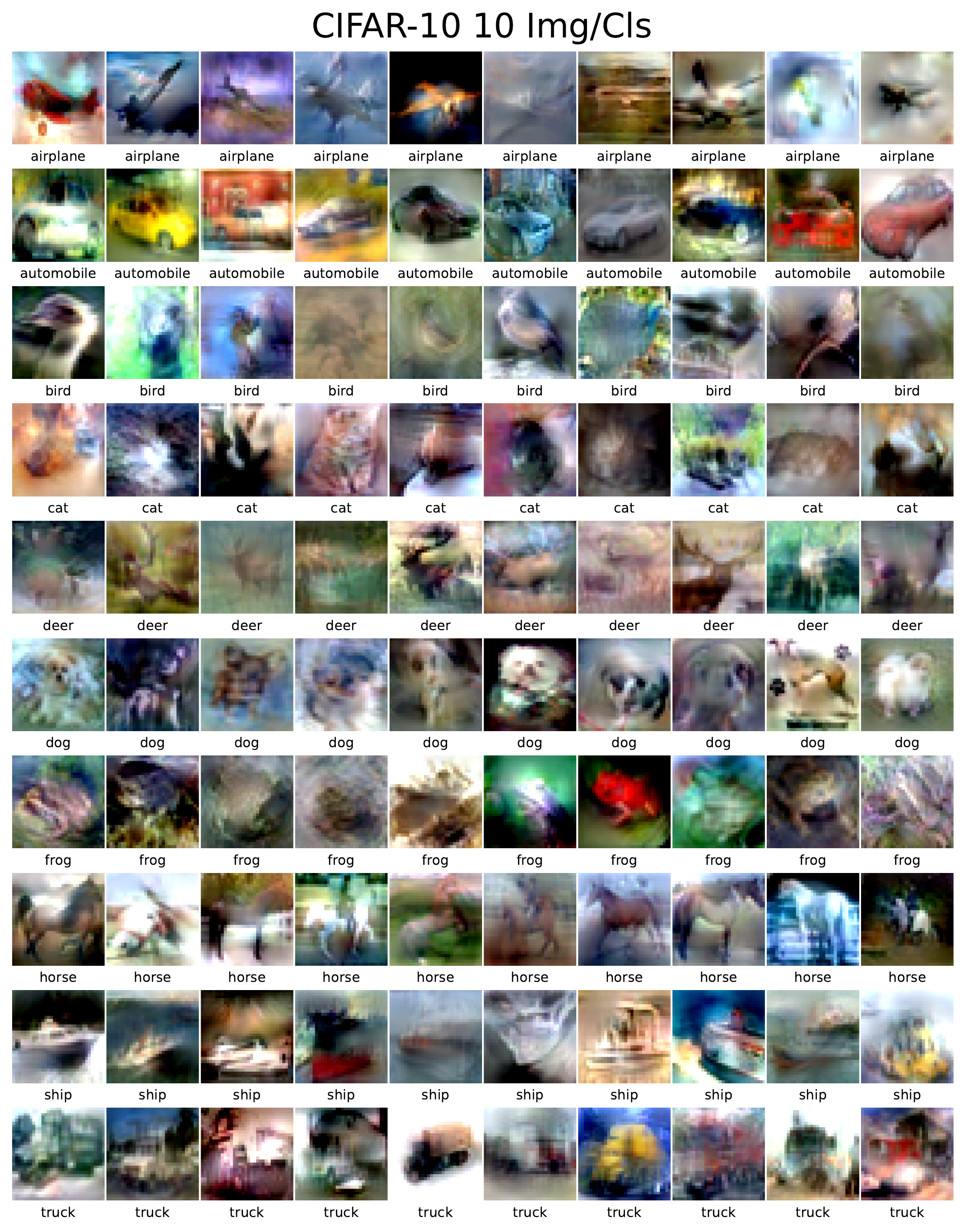}
\caption{RCIG distilled dataset on CIFAR-10 with 10 Img/Cls}
\end{center}
\vskip -0.2in
\end{figure*}

\begin{figure*}[h]
\vskip 0.1in
\begin{center}
\includegraphics[width = 0.7 \linewidth]{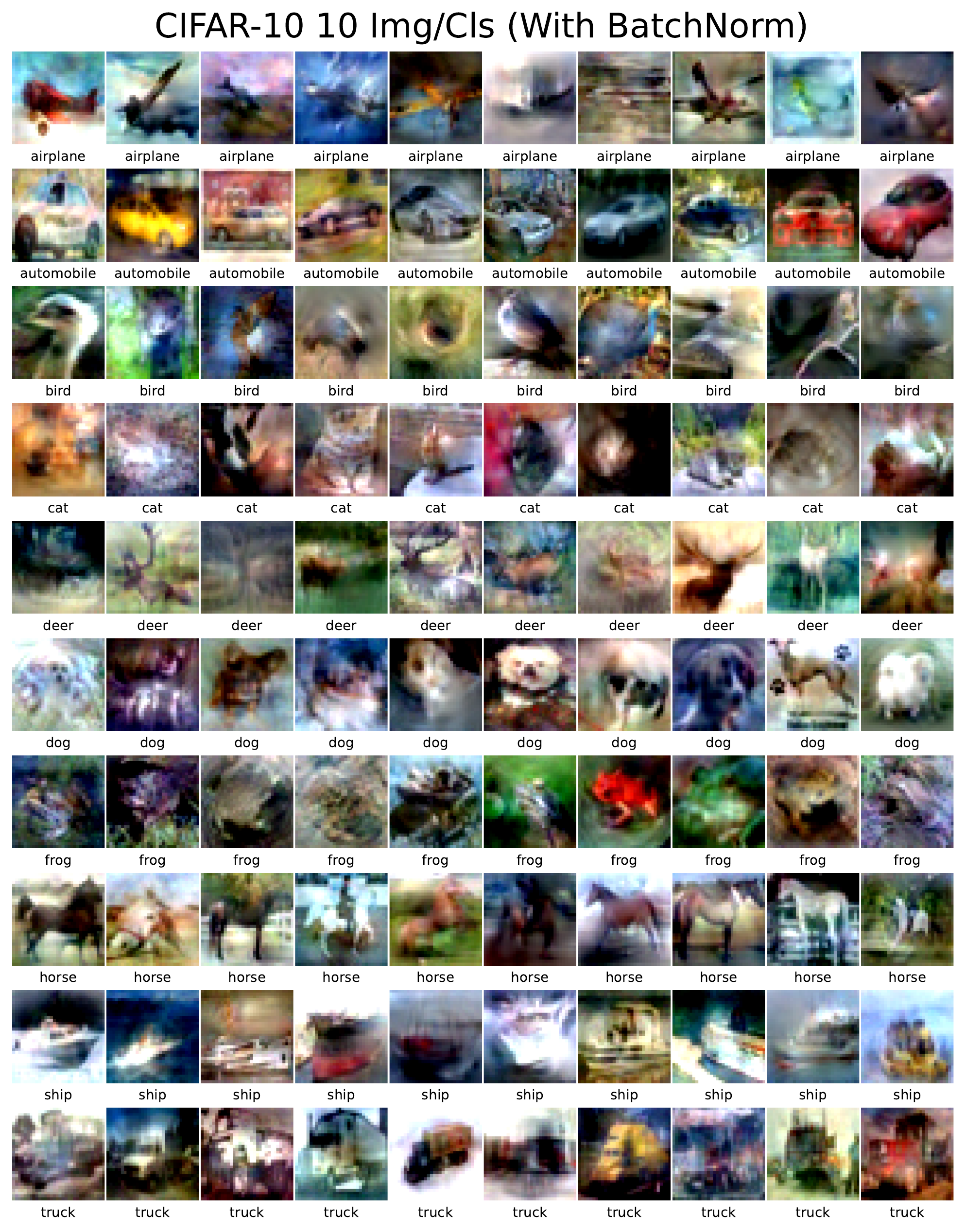}
\caption{RCIG distilled dataset on CIFAR-10 with 10 Img/Cls (Trained using BatchNorm)}
\end{center}
\vskip -0.2in
\end{figure*}

\begin{figure*}[h]
\vskip 0.1in
\begin{center}
\includegraphics[width = 0.7 \linewidth]{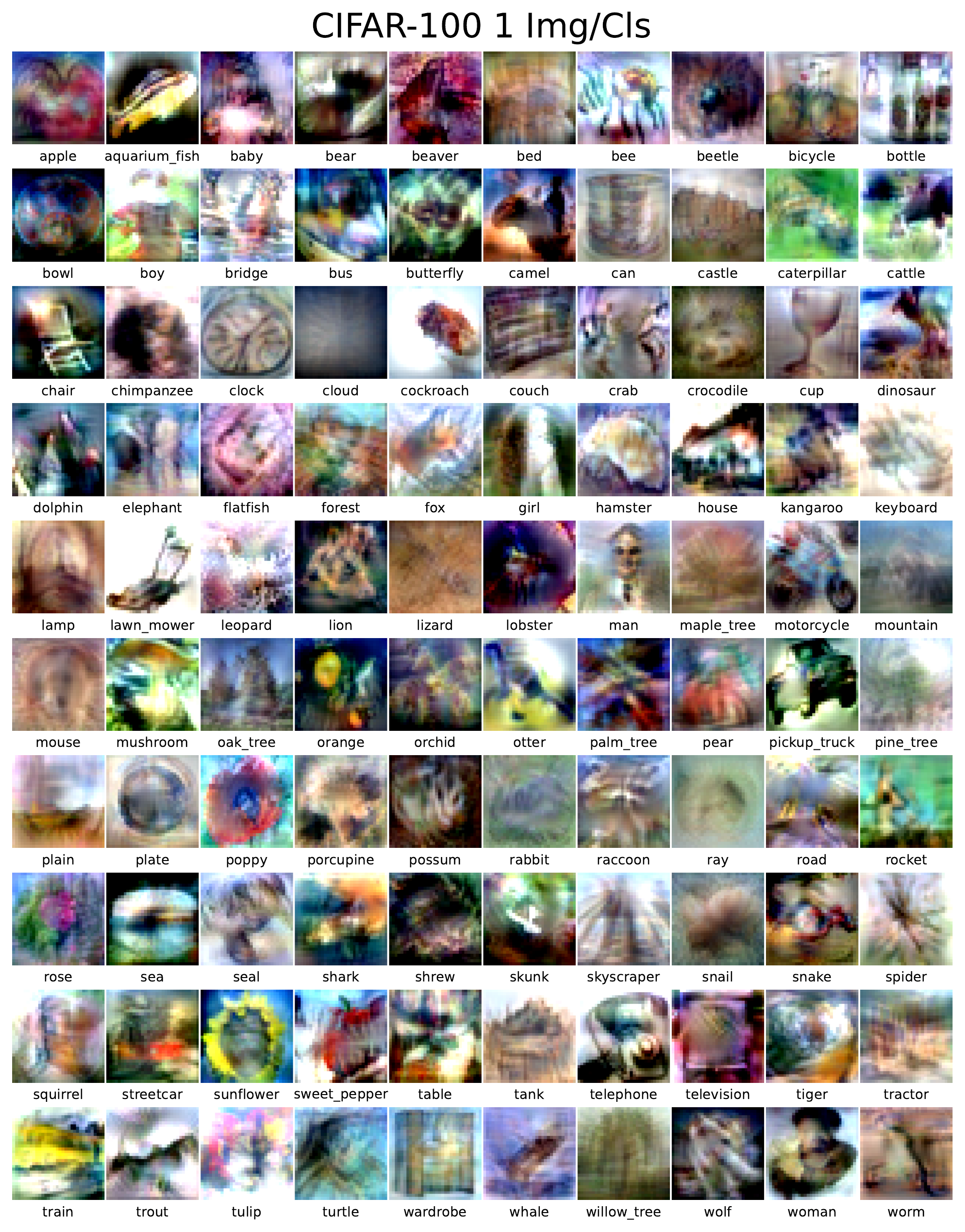}
\caption{RCIG distilled dataset on CIFAR-100 with 1 Img/Cls}
\end{center}
\vskip -0.2in
\end{figure*}

\begin{figure*}[h]
\vskip 0.1in
\begin{center}
\includegraphics[width = 0.5 \linewidth]{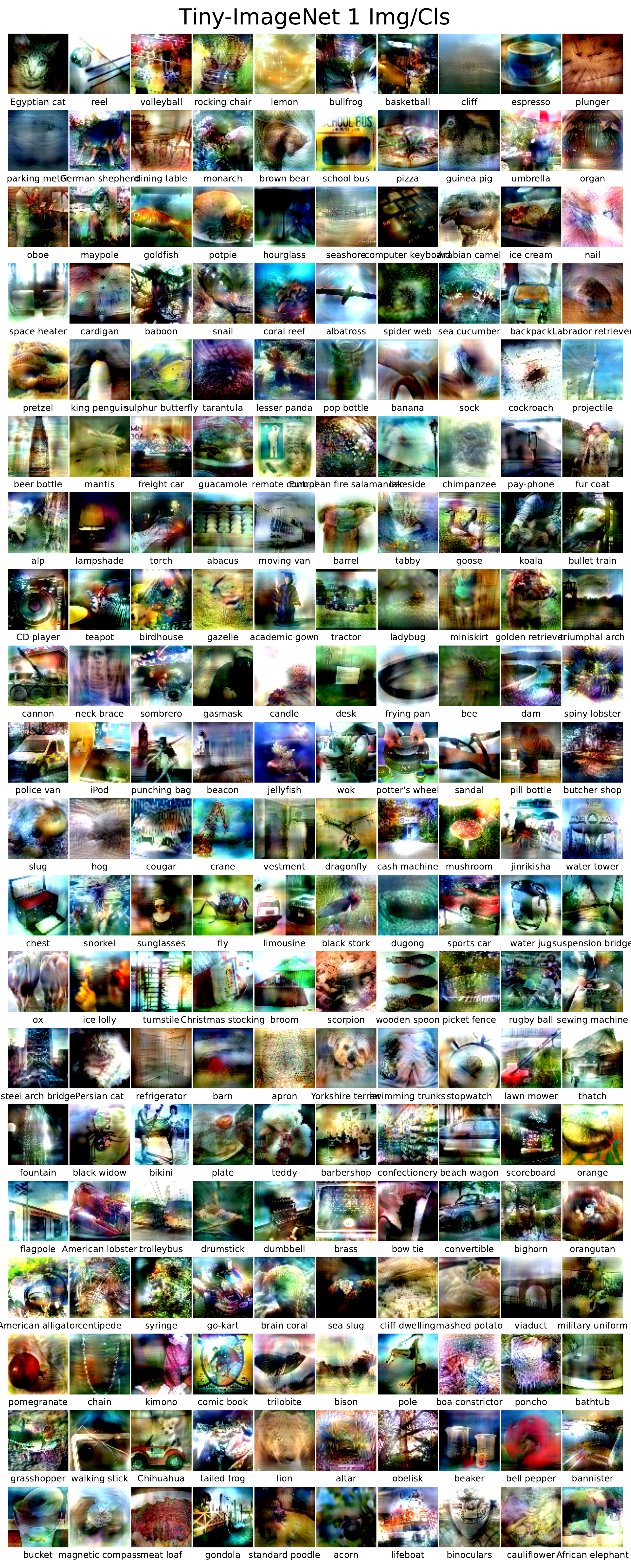}
\caption{RCIG distilled dataset on Tiny-ImageNet with 1 Img/Cls}
\end{center}
\vskip -0.2in
\end{figure*}

\begin{figure*}[h]
\vskip 0.1in
\begin{center}
\includegraphics[width = 0.7 \linewidth]{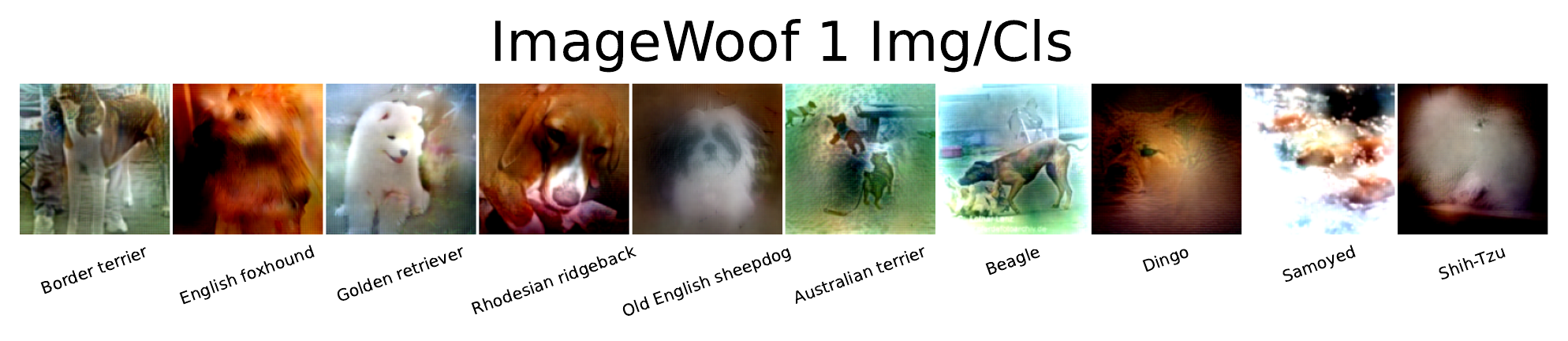}
\caption{RCIG distilled dataset on ImageWoof with 1 Img/Cls}
\end{center}
\vskip -0.2in
\end{figure*}

\begin{figure*}[h]
\vskip 0.1in
\begin{center}
\includegraphics[width = 0.7 \linewidth]{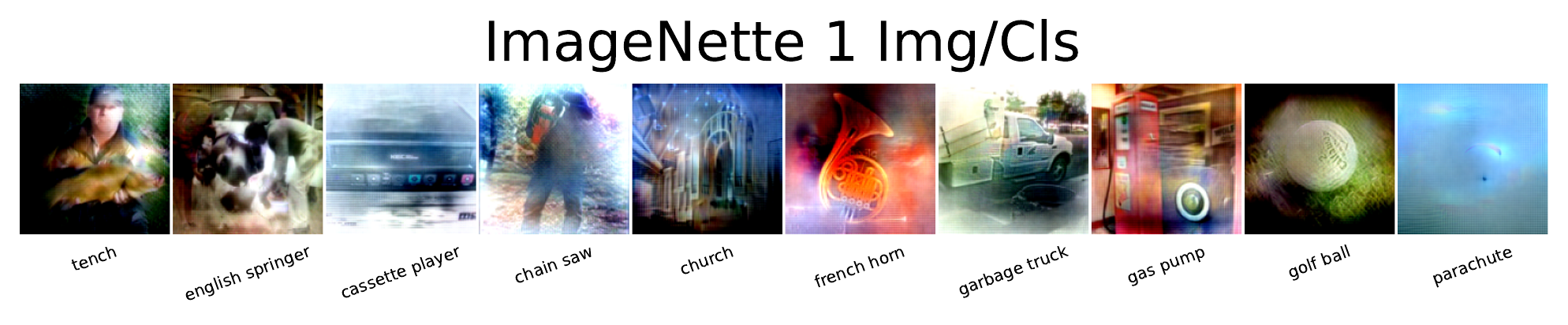}
\caption{RCIG distilled dataset on ImageNette with 1 Img/Cls}
\end{center}
\vskip -0.2in
\end{figure*}


\end{document}